\documentclass[runningheads]{llncs}
\usepackage{graphicx}

\usepackage{tikz}
\usepackage{comment}
\usepackage{amsmath,amssymb} 
\usepackage{color}

\usepackage[accsupp]{axessibility}

\usepackage{graphicx}
\usepackage{amsmath}
\usepackage{amssymb}
\usepackage{booktabs}
\usepackage{multirow}
\usepackage{color}
\usepackage{array}
\usepackage{float}
\usepackage{bbm}
\usepackage{makecell}
\usepackage[pagebackref=true,breaklinks=true,colorlinks,bookmarks=false]{hyperref}

\makeatletter
\newcommand\figcaption{\def\@captype{figure}\caption}
\newcommand\tabcaption{\def\@captype{table}\caption}
\makeatother

\usepackage[accsupp]{axessibility}  

\begin{document}
\pagestyle{headings}
\mainmatter

\title{Sim-to-Real 6D Object Pose Estimation via Iterative Self-training for Robotic Bin Picking}


\titlerunning{Iterative Self-training for Object Pose Estimation}
%

\author{Kai Chen\inst{1} \and
Rui Cao\inst{1} \and 
Stephen James\inst{2} \and
Yichuan Li\inst{1} \and\\
Yun-Hui Liu\inst{1} \and
Pieter Abbeel\inst{2} \and
Qi Dou\inst{1}}
\institute{The Chinese University of Hong Kong \and
University of California, Berkeley\\}
\authorrunning{K. Chen et al.}

\maketitle

\begin{abstract}
6D object pose estimation is important for robotic bin-picking, and serves as a prerequisite for many downstream industrial applications. However, it is burdensome to annotate a customized dataset associated with each specific bin-picking scenario for training pose estimation models.
In this paper, we propose an iterative self-training framework for sim-to-real 6D object pose estimation to facilitate cost-effective robotic grasping.
Given a bin-picking scenario, we establish a photo-realistic simulator to synthesize abundant virtual data, and use this to train an initial pose estimation network. This network then takes the role of a teacher model, which generates pose predictions for unlabeled real data. 
With these predictions, we further design a comprehensive adaptive selection scheme to distinguish reliable results, and leverage them as pseudo labels to update a student model for pose estimation on real data.
To continuously improve the quality of pseudo labels, we iterate the above steps by taking the trained student model as a new teacher and re-label real data using the refined teacher model. 
We evaluate our method on a public benchmark and our newly-released dataset, achieving an ADD(-S) improvement of 11.49\% and 22.62\% respectively.
Our method is also able to improve robotic bin-picking success by 19.54\%, demonstrating the potential of iterative sim-to-real solutions for robotic applications. Project homepage: \url{www.cse.cuhk.edu.hk/~kaichen/sim2real_pose.html}.
\end{abstract}

\section{Introduction}\label{sec:introduction}
6D object pose estimation aims to identify the position and orientation of objects in a given environment.
It is a core task for robotic bin-picking and plays an increasingly important role in elevating level of automation and reducing production cost for various industrial applications~\cite{drost2017introducing,kleeberger2019large,xianzhi2022s2r,wada2022safepicking,wada2022reorient}.
Though recent progress has been made, this task still remains highly challenging,
given that industrial objects are small and always cluttered together, leading to heavy occlusions and incomplete point clouds.
Early methods have relied on domain-specific knowledge and designed descriptors based on material textures or shapes of industrial objects~\cite{drost2010model,hinterstoisser2012model}.
Recently, learning-based models~\cite{he2020pvn3d,peng2020pvnet,tian2020shape,wada2020morefusion} that directly regress 6D pose from RGB-D inputs have emerged as more general solutions with promising performance for scalable application in industry.

However, training deep networks requires a large amount of annotated data. 
Different from labeling for ordinary computer vision
tasks such as object detection or segmentation, annotations of 6D object pose is extremely labor-intensive~(if possible), as it manages RGB-D data and involves several cumbersome steps, such as camera localization, scene reconstruction, and point cloud registration~\cite{brachmann2021visual,gojcic2020learning,murez2020atlas}. Without an easy way to get pose labels for real environments, the practicality of training object pose estimation networks lessens.

Perhaps then, the power of simulation can be leveraged to virtually generate data with labels of object 6D pose to train deep learning models for this task~\cite{hodan2018bop}. However, due to the simulation-to-reality gap, if we exclusively train a model with synthetic data, it is not guaranteed to also work well on real data. Sim-to-real adaptation methods are therefore required.
Some methods use physical-based simulation~\cite{hodavn2020bop} or mixed reality~\cite{wang2019normalized} to make simulators as realistic as possible. 
Other works~\cite{sundermeyer2020multi,sundermeyer2020augmented,tremblay2018deep} use domain randomization~\cite{james2017transferring,tobin2017domain} to vary the simulated environments with different textures, illuminations or reflections. 
Some recent methods resort to features that are less affected by the sim-to-real gap for object pose estimation~\cite{georgakis2019learning,li2021sd}.
Alternatively, some methods initially train a model on synthetic data, and then refine the model by self-supervised learning on unlabeled real data~\cite{deng2020self,wang2020self6d}.
Unfortunately, these sim-to-real methods either rely on a customized simulation to produce high-quality synthetic data or need a customized network architecture for extracting domain-invariant signal for sim-to-real pose estimation. Given that a robot would be applied to various upcoming bin-picking scenarios, customizing the method to fit for various different real scenes could also be labor-intensive. In addition, since the adaptation to complex bin-picking scenarios is not sufficient, obvious performance gap between simulation and real evaluation is still present in existing sim-to-real methods.

These current limitations encourage us resorting to a new perspective to tackle this challenging task.
In general, a model trained on synthetic data should have acquired essential knowledge about how to estimate object poses. If we directly expose it to real data, its prediction is not outrageously wrong. If the model could directly self-train itself leveraging the successful practices, it is highly likely to efficiently adapt to reality, just with unlabeled data. Collecting a large amount of unlabeled real data is easy to achieve for a robot. More importantly, this adaptation process is scalable and network-agnostic. In any circumstances, the robot can be deployed with the most suitable deep learning network for its target task, record sufficient unlabeled real data, and then leverage the above self-training adaptation idea for accurate object pose estimation in its own scenario. 
Promisingly, similar self-training paradigms have been recently explored for some other computer vision applications such as self-driving and medical image diagnosis, for handling distribution shift towards real-world use~\cite{pastore2021closer,roychowdhury2019automatic,wang2021deep}. 
However, the question of how it can be used for reducing annotation cost while keeping up real-data performance of 6D object pose estimation for robots is unknown.

\begin{figure}[t]
    \centering
    \includegraphics[width=0.95\linewidth]{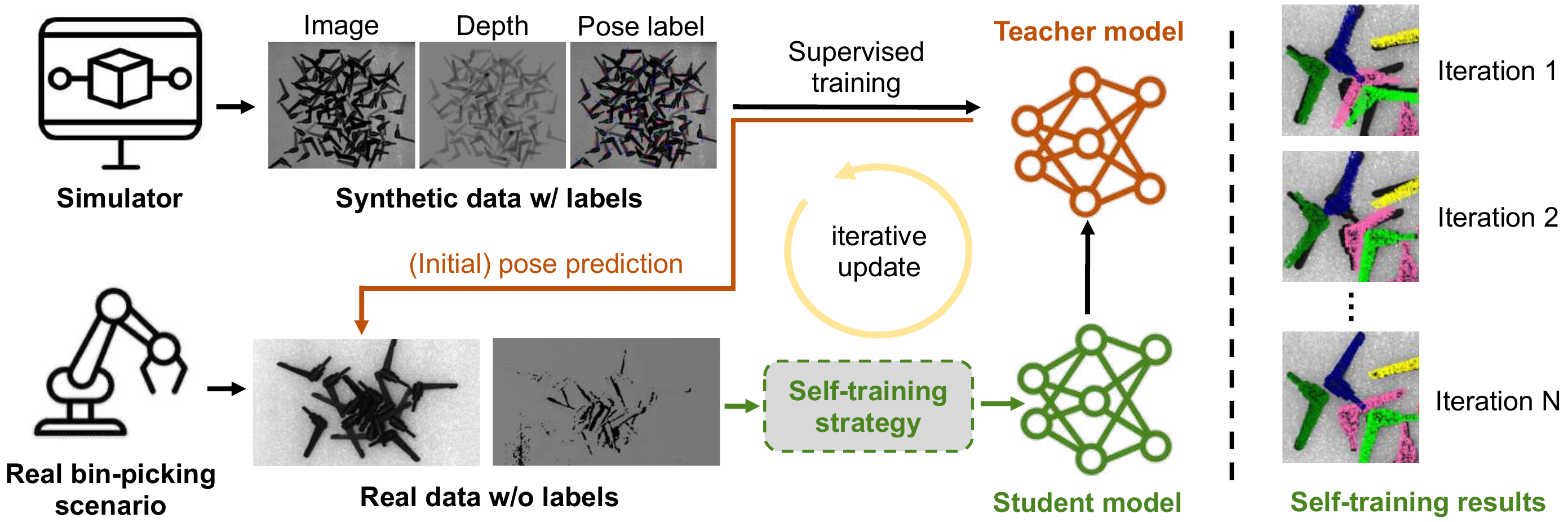}
    \caption{For industrial bin-picking, we build a simulator to generate synthetic data for learning a teacher model. Starting from it, our novel iterative self-training method can select reliable pseudo-labels and progressively improve the 6D object pose accuracy on real data, without the need for any manual annotation.}
    \label{fig:cover_page}
\end{figure}

In this paper, we propose the first sim-to-real iterative self-training framework for 6D object pose estimation.
Fig.~\ref{fig:cover_page} depicts an overview of our framework. First, we establish a photo-realistic simulator with multiple rendering attributes to synthesize abundant data, on which a pose estimation network is initially well-trained. Then, it is taken as a teacher model to generate pose predictions for unlabeled real data. 
From these results, we design a new pose selection method, which comprehensively harnesses both 2D appearance and 3D geometry information of objects to carefully distinguish reliable predictions.
These high-quality pose predictions are then leveraged as pseudo labels for the corresponding real data to train the pose estimation network as a student model.
Moreover, to progressively improve the quality of pseudo labels, we consider the updated student model as a new teacher, refine pose predictions, and re-select the highest-quality pseudo labels again. 
Such an iterative self-training scheme ensures to simultaneously improve the quality of pseudo labels and model performance on real data. According to extensive experiments on a public benchmark dataset and our constructed \emph{Sim-to-real Industrial Bin-Picking~(SIBP)} dataset, our proposed iterative self-training method significantly outperforms the state-of-the-art sim-to-real method~\cite{sundermeyer2020augmented} by 11.49\% and 22.62\% in terms of the ADD(-S) metric.

\section{Related Works}\label{sec:related_works}
\subsection{6D Object Pose Estimation for Bin-Picking}\label{subsec:rw_pose} 
Though recent works~\cite{di2021so,hodavn2020bop,peng2020pvnet,sundermeyer2020multi} show superior performance on household object datasets (\textit{e.g.,} LineMOD~\cite{hinterstoisser2012model} and YCB Video~\cite{xiang2017posecnn}), 6D pose estimation for bin-picking still remains a challenging task for highly cluttered and complex scenes. Conventional methods typically leveraged Point Pair Features~(PPF) to estimate 6D object pose in a bin~\cite{buch2017rotational,buch2018local,drost2010model,tuzel2014learning}. As a popular method, it mainly relies on depth or point cloud to detect the pose in the scene and achieves promising results when scenes are clear and controlled~\cite{hodavn2020bop}. However, due to the reliance on the point cloud, the PPF-based methods are sensitive to the noise and occlusion in cluttered scenes, which is very common in real bin-picking scenarios such as industrial applications~\cite{kleeberger2020single,yang2021robi}.
Current state-of-the-art methods~\cite{dong2019ppr,kleeberger2020single} and datasets~\cite{kleeberger2019large,yang2021robi} rely on deep learning models for handling complex bin-picking scenarios. Dong et al.~\cite{dong2019ppr} proposed a deep-learning-based voting scheme to estimate 6D pose using a scene point cloud.  Yang et al.~\cite{yang2021robi} modified AAE~\cite{sundermeyer2018implicit} with a detector as a deep learning baseline, showing a better performance than the PPF-based method for bin-picking scenes. For these methods, regardless of the particular network design, a large amount of real-world labeled data is always required in order to achieve a high accuracy.

\subsection{Visual Adaptation for Sim-to-real Gap}\label{subsec:rw_sim_to_real}
Sim-to-real transfer is a crucial topic to tackle the bottleneck for deep learning-based robotic applications, which is essential to bridge the domain gap between simulated data and real-world observation through robotic visual systems~\cite{manhardt2018deep,wang2020self6d}. 
A promising technique for achieving sim-to-real transfer is domain randomization~\cite{kehl2017ssd,manhardt2018deep,tremblay2018deep}, which samples a huge diversity of simulation settings (e.g. camera position, lighting, background texture, etc.) in a simulator to force the trained model to learn domain-invariant attributes and to enhance generalization on real-world data. Visual domain adaptation~\cite{sundermeyer2020multi,sundermeyer2018implicit,sundermeyer2020augmented} has also had recent success for sim-to-real transfer, where the source domain is usually the simulation, and the target domain is the real world~\cite{bousmalis2018using,james2019sim}. Recently, some works~\cite{di2021so,hodavn2020bop,tremblay2018falling,tremblay2018deep,wang2020self6d} leverage physically-based renderer~(PBR) data to narrow the sim-to-real gap by simulating more realistic lighting and texture in the simulator. However, with the additional synthetic data, current works on 6D pose estimation still mainly rely on annotated real data to ensure performance.

\subsection{Self-training via Iterative Update}\label{subsec:rw_self_training}
A typical self-training paradigm~\cite{grandvalet2004semi,yarowsky1995unsupervised} first trains a teacher model based on the labeled data, then utilizes the teacher model to generate pseudo labels for unlabeled data. After that, the unlabeled data with their pseudo labels are used to train a student model. As a typical label-efficient methodology, self-training techniques have recently gained attention in many fields. Some methods apply self-training to image classification~\cite{veit2017learning,yalniz2019billion}. They use the teacher model to generate soft pseudo labels for unlabeled images, and train the student model with a standard cross-entropy loss. The input consistency regularization~\cite{xie2020unsupervised} and noisy student training~\cite{xie2020self} are widely used techniques to further enhance the student model with unlabeled data. Some works adopt self-training in semantic segmentation~\cite{pastore2021closer,zhu2020improving}. Instead of training the student model with merely pseudo labels, they train the student by jointly using a large amount of data with pseudo labels and a small number of data with manual labels. Some other works resort to self-training to perform unsupervised domain adaptation~\cite{roychowdhury2019automatic,zou2018unsupervised}, which train the teacher and student models on different domains. Recently, some methods study pseudo label selection. They pick out reliable pseudo labels based on the probability outputs or the uncertainty measurements derived from the network~\cite{gal2016dropout,rizve2020defense}. Though the effectiveness of self-training has been revealed in many fields, we find that self-training has not been investigated for 6D object pose estimation.

\section{Method}\label{sec:method}
\begin{figure*}[t]
    \centering
    \includegraphics[width=0.9\textwidth]{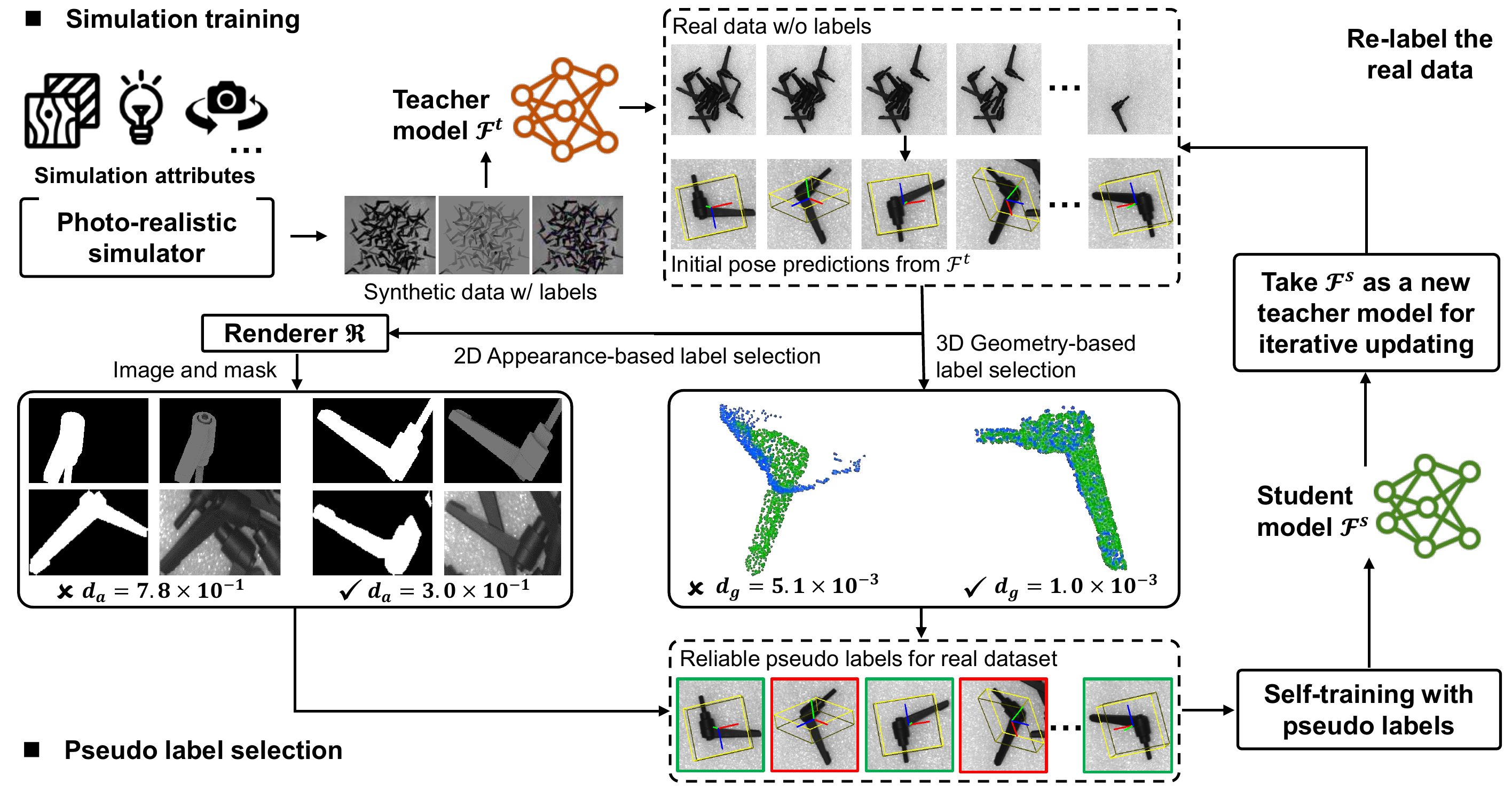}
    \caption{Overview of our proposed iterative self-training method for 6D object pose estimation. We first build a photo-realistic simulation with multiple rendering attributes to generate synthetic data. We then train a teacher model on the synthetic data, which is used to generate initial pose predictions for the unlabeled real data. Subsequently, a pseudo label selection scheme with both 2D appearance and 3D geometry metrics are adopted to select data with reliable pose predictions, which are used to train a student model for the real data. We iterate this self-training scheme multiple times by taking the trained student model as a new teacher for the next iteration.}
    \label{fig:method_overview}
\end{figure*}
In this paper, we take advantage of advanced self-training to solve the problem of sim-to-real object pose estimation. Our method is designed to be scalable and network-agnostic, and aims to dramatically improve pose estimation accuracy on real-world data without need of any manual annotation.

\subsection{Overview of Sim-to-Real Object Pose Estimation}\label{subsec:overview}
Without loss of generality, let $\mathcal{F}$ denote an arbitrary 6D object pose estimation network, which takes RGB-D data of an object as input, and outputs its corresponding 6D pose with respect to the camera coordinate frame. Let $I$ be the RGB image and $G$ denote the 3D point cloud recovered from the depth map, the 6D object pose $p$ is predicted by $\mathcal{F}$ as:
\begin{equation}
    p=[R|t]=\mathcal{F}(I, G),
\end{equation}
where $R\in SO(3)$ is the rotation and $t\in\mathbb{R}^3$ is the translation. Fig.~\ref{fig:method_overview} summarizes our network-agnostic, sim-to-real pose estimation solution. Given a target bin-picking scenario, we first create a photo-realistic simulation to generate abundant synthetic data with a low cost, which is used to initially train a teacher model $\mathcal{F}^t$ with decent quality.
We then apply $\mathcal{F}^t$ on real data to predict their object poses~(Sec.~\ref{subsec:label_generation}). Based on these initial estimations, a new robust label selection scheme~(Sec.~\ref{subsec:label_selection}) is designed to select the most reliable predictions for pseudo labels, which are incorporated by self-training to get a student model $\mathcal{F}^s$ for the real data~(Sec.~\ref{subsec:self-training}). Importantly, we iterate the above steps by taking the trained $\mathcal{F}^s$ as a new teacher model, in order to progressively leverage the knowledge learned from unlabeled real data to boost the quality of pseudo labels as well as the student model performance on real data.

\subsection{Simulation Training}\label{subsec:label_generation}
First of all, we construct a photo-realistic simulator to generate synthetic data for the bin-picking scenario. To mimic a real-world cluttered bin-picking scenario, we randomly generate realistic scenes with multiple objects closely stacked in diverse poses. 
Note that our simulator is light-weight without the need for complex scene-orientated modules~(such as mixed/augmented reality), which means that once built, the simulator can be widely applied to different industrial bin-picking scenarios. Using the simulator, we generate a large amount of data with precise labels, and use it to train a teacher model $\mathcal{F}^t$ with acceptable initial performance. After that, when we expose $\mathcal{F}^t$ to real data $(I_i, G_i)$, we can obtain its pose prediction as $\widetilde{p_i}=\mathcal{F}^t(I_i, G_i)$. As shown in Fig.~\ref{fig:method_overview}, due to the unavoidable gap between virtual and real data, $\widetilde{p_i}$ is not guaranteed to always be correct. In order to self-train the model on unlabeled real data, we need to carefully pick out reliable predictions of $\mathcal{F}^t$. This step is known as pseudo label selection in a typical self-training paradigm.

\subsection{Pose Selection for Self-training}\label{subsec:label_selection}
\begin{figure*}[t]
    \centering
    \includegraphics[width=1.0\linewidth]{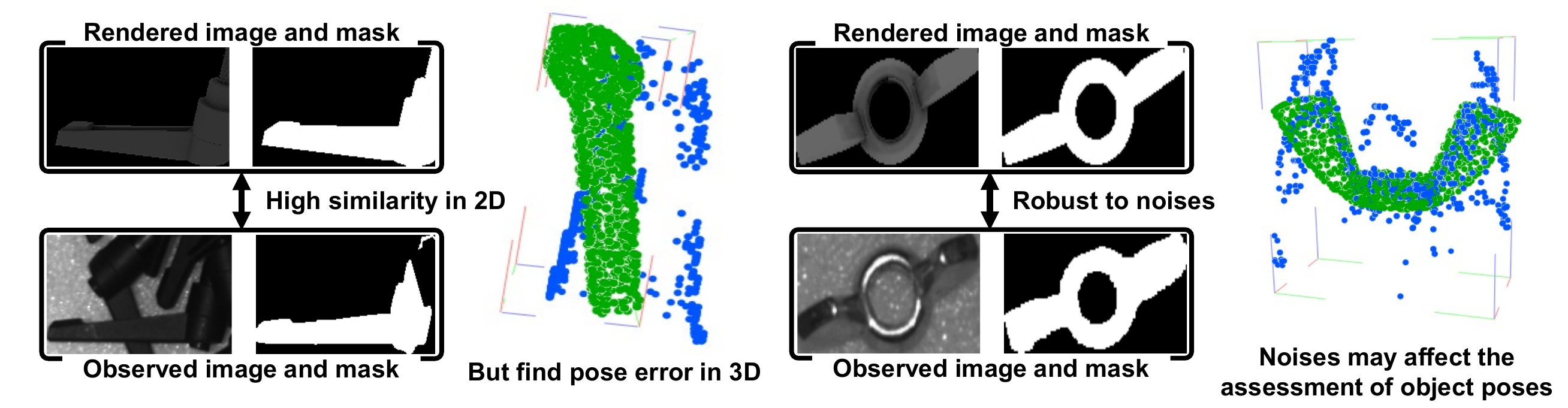}
    \caption{Both 2D appearance and 3D geometry are important for reliable pseudo label selection. On the one hand, the 3D geometry metric can notice pose errors that are hard to detect in 2D. On the other hand, 3D geometry metric gets unreliable for noisy point cloud, but the 2D appearance metric is robust to these noises.}
    \label{fig:2d3d_pseudo_label_selection}
\end{figure*}

Given unlabeled real data $\{(I_i, G_i)\}_{i=1}^m$ and their initial pose predictions $\{\widetilde{p}_i=[R_i|t_i]\}_{i=1}^m$, pose selection aims to find out reliable pose results.
Existing pseudo label selection methods~\cite{gal2016dropout,rizve2020defense,xie2020self} are limited to classification-based tasks, such as image recognition and semantic segmentation. But for object pose estimation, which is typically formulated as a regression learning process, it is unclear how to select pseudo labels. In order to address this problem, we propose a comprehensive pose selection method. The core idea is to first virtually generate the observation data that corresponds to the predicted pose. Subsequently, we compare the generated data with the real collected one to determine whether the predicted pose is reliable or not. Note that for a robust pose selection, both the 2D image and the 3D point cloud are important. As shown in Fig.~\ref{fig:2d3d_pseudo_label_selection}, although 2D image provides rich features for discriminating object poses, they could be ambiguous due to the information loss when projecting the 3D object onto the 2D image plane. Using 3D point cloud to assess the object pose can avoid this issue. But different from 2D images, the 3D point cloud could be quite noisy in the bin-picking scenario, which affects the assessment result. Our proposed method therefore leverages the complementary advantages of 2D image and 3D point cloud for a comprehensive pose selection for self-training.

\subsubsection{2D Appearance-based Pose Selection.} To obtain the image that corresponds to the predicted pose, we leverage an off-the-shelf renderer~\cite{chen2019learning}, which is denoted as $\Re$. The renderer takes an object pose and the corresponding object CAD model as inputs, and will virtually generate an RGB image $I^r$ and a binary object mask $M^r$ that correspond to the predicted pose\footnote{For clarity, we will omit the subscript $i$ for formulations in this section.}:
\begin{equation}
    \{I^r, M^r\}=\Re(\widetilde{p}).
\end{equation}
The binary mask exhibits clear object contour, while the RGB image contains richer texture and semantic information (see Fig.~\ref{fig:method_overview}). We make combined use of them for appearance-based pose selection on the 2D image plane.

Let $M^o$ denote the object mask for the observed RGB image, which can be accurately acquired by an existing segmentation network~\cite{he2017mask} trained on synthetic data. Rather than directly measuring the contour similarity~\cite{manhardt2018deep} and to ensure that the mask-based metric is robust to segmentation noise, we evaluate the pose quality based on the pixel-wise overlap of $M^r$ and $M^o$, which is less affected by imperfect segmentation:
\begin{equation}
    s(M^r, M^o)=\frac{1}{2}\times(\frac{1}{N_+}\sum_{q\in{M^o_+}}\mathbbm{1}(q\in{M^r_+})+\frac{1}{N_{-}}\sum_{q\in{M^r_-}}\mathbbm{1}(q\in{M^o_-})),
\end{equation}
where $M_+^{\{r,o\}}$ denotes the foreground region of the mask and $M_-^{\{r,o\}}$ denotes the background region of the mask. $N_+$ and $N_-$ are the number of pixels of $M_+^o$ and $M_-^r$, respectively. Given the mask overlap, the corresponding mask-based pose distance metric could be computed as $d_{mask}=1-s(M^r, M^o)$. This mask-based metric coarsely leverages the object contour information for pose evaluation, but neglects the detailed texture and semantic information of the object. As a consequence, $d_{mask}$ would be sensitive to occlusion and not reliable enough for objects with a complex shapes.

In this regard, we further leverage the rendered RGB image to enhance the pose assessment. In order to extract representative features from the RGB image, we use a pre-trained CNN $\Phi$ to transform $I^r$ and $I^o$ into multi-level high-dimension features, based on which we measure the perceptual distance~\cite{zhang2018unreasonable} between $I^r$ and $I^o$ as:
\begin{equation}
    d_{image}=\sum_{l=1}^L\frac{1}{N_l}\sum_q\|\Phi_l^r(q)-\Phi_l^o(q)\|_2,
\end{equation}
where $\Phi_l^{\{r,o\}}$ denotes the $l$-th level normalized feature of $\Phi$. $N_l$ is the number of pixels of the corresponding feature map. The perceptual distance $d_{image}$ complements $d_{mask}$ with low-level texture and high-level semantic features. We then integrate them as $d_{a}=d_{mask}\times d_{image}$ to assess the initial pose prediction $\widetilde{p}$ on the 2D image plane.

\subsubsection{3D Geometry-based Pose Selection.} The 2D appearance-based metric is good at measuring the in-plane translation and rotation with informative features on 2D image plane. Nevertheless, it is not sufficient to comprehensively assess a complete 6D pose.
To address this limitation, we propose to further leverage the object point cloud $G^o$ to enhance the pose selection scheme in 3D space.
In order to use the point cloud consistency in 3D space to assess the pose quality, we need to generate the object point cloud that corresponds to the predicted pose. With the object CAD model $\mathbb{C}$ and the predicted pose $\widetilde{p}$, we perform the following 2 steps. First, we apply $\widetilde{p}$ to the CAD model: $\mathbb{C}^r=\widetilde{R}\times\mathbb{C}+\widetilde{t}$. It transforms the original CAD model $\mathbb{C}$ in the object coordinate frame to a model $\mathbb{C}^r$ in the camera coordinate frame. After that, since $\mathbb{C}^r$ is a complete point cloud while $G^o$ only contains the surface point cloud that could be observed from a view point, directly evaluating the point cloud consistency between $\mathbb{C}^r$ and $G^o$ cannot precisely reveal the quality of $\widetilde{p}$. To mitigate this problem, we further apply a projection operator proposed by~\cite{gu2020weakly} to the CAD model in camera coordinate frame: $G^r=\mathbb{P}(\mathbb{C}^r)$. This generates the surface points of $\mathbb{C}^r$ that can be observed from the predicted pose. Intuitively, if $\widetilde{p}$ is precise, $G^r$ should be well aligned with the observed point cloud $G^o$ in 3D space. We use the Chamfer Distance~(CD) between $G^r$ and $G^o$ as a 3D metric to quantify the quality of $\widetilde{p}$:
\begin{equation}\label{eq:r}
    d_{g}=\frac{1}{N_1}\sum_{x\in G^r}\min_{y\in G^o}\|x-y\|_2 + \frac{1}{N_2}\sum_{y\in G^o}\min_{x\in G^r}\|x-y\|_2,
\end{equation}
where $x$ and $y$ denote 3D points from $G^r$ and $G^o$, respectively. $N_1$ and $N_2$ are the total numbers of points for $G^r$ and $G^o$.

\subsubsection{Comprehensive Selection.} Given the complementary advantages of $d_a$ and $d_g$, we leverage both of them to single out reliable pose predictions. Pose predictions are designated as pseudo labels and included in the next iteration of self-training when $d_a < \tau_a$ and $d_g < \tau_g$. These two thresholds $\tau_a$ and $\tau_g$ can be determined flexibly according to the metric value distribution on unlabeled real data. In this paper, we adaptively set $\tau_{(a,g)}=\mu_{(a,g)}+\sigma_{(a,g)}$, where $\mu_{(a,g)}$ and $\sigma_{(a,g)}$ are the mean and standard deviation of $d_a$ and $d_g$ on real data.

\subsection{Iterative Self-training on Real Data}\label{subsec:self-training}
After label selection, we incorporate data with reliable pseudo labels to the 6D object pose estimation network to train a student model $\mathcal{F}^s$. Our self-training scheme aims to enforce the prediction of $\mathcal{F}^s$ to be consistent with pose pseudo labels. In this regard, we define the self-training loss function as $\mathcal{L}_\text{self-training}=\frac{1}{M}\sum_{i=1}^M\ell(\widetilde{p}_i, p_i'),$
where $p_i'$ is the pose prediction from the student model, $\ell$ could be any loss function of object pose estimation and is depended on the architecture of $\mathcal{F}$. Through self-training, we transfer the knowledge that the teacher model has learned from the synthetic data to the student model on real data. 

To further improve the capability of the student model, we iterate the above process for multiple runs. With progressive training on the real data, the pseudo labels generated by the updated teacher model will have a higher quality than the previous iteration's labels. Our experiments demonstrate that this iterative optimization scheme can effectively boost the student model on real data.
\section{Experiments}\label{sec:experiment}
In this section, we will answer the following questions: (1) Does the proposed self-training method improve the performance of sim-to-real object pose estimation? (2) Do the appearance-based and geometry-based pose selection metrics improve the student model performance? (3) Could the proposed self-training framework be applied to different backbone architectures? (4) How many iterations are required to self-train the pose estimation network? (5) Does the proposed self-training indeed improve the bin-picking success rate on real robots? To answer question (1)-(4), we evaluate our method and compare it with state-of-the-art sim-to-real methods on both the ROBI~\cite{yang2021robi} dataset and our proposed SIBP dataset. To answer question (5), we deploy our pose estimation model after self-training on a real robot arm and conduct real-world bin-picking experiments.

\subsection{Experiment Dataset}
\textbf{ROBI Dataset.} ROBI~\cite{yang2021robi} is a recent public dataset for industrial bin-picking. It contains seven reflective metal objects, with two of them being composed of different materials\footnote{Din-Connector and D-Sub Connector}. Since ROBI only contains real RGB-D data, we use the simulator presented in Sec.~\ref{subsec:label_generation} to extend ROBI by randomly synthesizing $1000$ virtual scenes for each object. For real data, we use 3129 real RGB-D scenes of ROBI. The 3129 real scenes are in two cluttered levels. 1288 scenes are low-bin scenarios, in which a bin contains only 5-10 objects and corresponds to easy cases. 1841 sceres are full-bin scenarios, in which a bin is full with tens of objects and correspond to hard cases. We use 2310 scenes~(910 low-bin and 1400 full-bin scenarios) without using their annotations for self-training and use the remaining scenes for evaluation. Please refer to~\cite{yang2021robi} and the supplementary for more detailed information of ROBI.\\

\noindent\textbf{Sim-to-real Industrial Bin-Picking~(SIBP) Dataset.} We further build a new SIBP dataset. It provides both virtual and real RGB-D data for six texture-less objects in industrial bin-picking scenarios. Compared with ROBI, objects in SIBP are more diverse in color, surface material~(3 plastic, 2 metallic and 1 alloyed) and object size~(from a few centimeters to one decimeter). We use SIBP for a more comprehensive evaluation. For synthetic data, we use the same simulator to generate 6000 synthetic RGB-D scenes. For real data, we
use a Smarteye Tech HD-1000 industrial stereo camera to collect 2743 RGB-D scenes in a real-world bin-picking environment. We use 2025 scenes without using their annotations for self-training, and use the remaining scenes for evaluation. Please refer to supplementary for more detailed information of SIBP.

\subsection{Evaluation Metrics}
We follow~\cite{sundermeyer2020multi,thalhammer2021pyrapose,wang2020self6d} and compare pose estimation results of different methods $w.r.t.$ the ADD metric~\cite{hinterstoisser2012model}, which measures whether the average distance between models transformed by the predicted pose and the ground-truth pose is smaller than $10\%$ of object diameter. For symmetric objects, we adopted the ADD-S metric~\cite{hodavn2016evaluation} when computing the average distance of models. We report the average recall of ADD(-S) for quantitative evaluation.


\subsection{Comparison with State-of-the-art Methods}
We compared our self-training model with a baseline DC-Net~\cite{tian2020robust} trained exclusively with synthetic data, and two state-of-the-art sim-to-real object pose estimation methods: MP-AAE~\cite{sundermeyer2020multi} and AAE~\cite{sundermeyer2020augmented}. Note that all methods are evaluated based on the same segmentation. Tab.~\ref{table:quantitative_evaluation_robi} and Tab.~\ref{table:quantitative_evaluation_sibp} presents the quantitative evaluation results. In comparison with the baseline model, on ROBI, our self-training model outperforms DC-Net by $16.73\%$. On SIBP, it also exceeds DC-Net by $14.94\%$. These results demonstrate the effectiveness of our self-training method for mitigating the sim-to-real gap for object pose estimation. 

Moreover, we compared our method with two SOTA sim-to-real methods. AAE first trained an Augmented Autoencoder on synthetic data and then used the embedded feature of Autoencoder for object pose estimation on real data. MP-AAE further resorted to multi-path learning to learn a shared embedding space for pose estimation of different objects on real data. 
Our self-training method consistently outperforms these two SOTA methods for object pose estimation. On ROBI, our model outperforms MP-AAE by $24.5\%$ and exceeds AAE by $11.49\%$. On SIBP, our self-training model achieves an average recall of $68.72\%$, which is $32.79\%$ higher than MP-AAE and $22.62\%$ higher than AAE. 
Fig.~\ref{fig:qualitative_comparison} presents corresponding qualitative comparisons on ROBI. MP-AAE and AAE have difficulties in adapting the model to a real bin-picking environment. The occlusions caused by the container or by neighboring objects may affect their pose estimation results. In comparison, our method can directly adapt the pose estimation model to real data with self-training. It can smoothly handle complex industrial bin-picking scenarios and achieve high pose estimation accuracy. Please refer to the supplementary for more qualitative results.

\begin{figure*}[t!]
	\centering
		\begin{minipage}{1.0\textwidth}
			\centering
			\tabcaption{Quantitative comparison with state-of-the-art methods on ROBI dataset.The average recall~(\%) of the ADD(-S) metric is reported. * indicates asymmetric objects.}
            \resizebox{1.0\textwidth}{!}{%
            \renewcommand\arraystretch{1.0}
            \begin{tabular}{l|ccccccc|c}
                 \toprule
                 Object & Zigzag* & \makecell[c]{Chrome\\Screw} & Gear & \makecell[c]{Eye\\Bolt} & \makecell[c]{Tube\\Fitting} & \makecell[c]{Din*\\Connector} & \makecell[c]{D-Sub*\\Connector} & {Mean}\\\hline
                 MP-AAE~\cite{sundermeyer2020multi} & {22.76} & {47.49} & {66.22} & {50.85} & {69.54} & {15.92} & {4.70} & {39.64}\\
                 AAE~\cite{sundermeyer2020augmented} & {25.80} & {64.02} & {90.64} & {67.31} & {91.58} & {21.83} & {7.35} & {52.65}\\
                 DC-Net~\cite{tian2020robust} & {30.96} & {67.74} & {77.56} & {53.52} & {74.84} & {18.72} & {10.60} & {47.41}\\
                 Ours & {\textbf{45.83}} & {\textbf{79.88}} & {\textbf{97.81}} & {\textbf{89.21}} & {\textbf{97.45}} & {\textbf{24.21}} & {\textbf{14.62}} & {\textbf{64.14}}\\
                 \bottomrule
            \end{tabular}}
            \label{table:quantitative_evaluation_robi}
		\end{minipage}
		\vfill
		\begin{minipage}[h]{1.0\linewidth}
			\centering
			\tabcaption{Quantitative comparison with state-of-the-art methods on SIBP dataset.The average recall~(\%) of the ADD(-S) metric is reported. * indicates asymmetric objects.}
            \resizebox{1.0\textwidth}{!}{%
            \renewcommand\arraystretch{1.0}
            \setlength{\tabcolsep}{1.1mm}{
            \begin{tabular}{l|cccccc|c}
                \toprule
                 Object & Cosmetic & Flake & Handle* & Corner & \makecell[c]{Screw\\Head} & \makecell[c]{T-Shape\\Connector} & {Mean}\\\hline
                MP-AAE~\cite{sundermeyer2020multi} & {58.04} & {21.56} & {27.26} & {39.59} & {36.67} & {32.44} & {35.93}\\
                 AAE~\cite{sundermeyer2020augmented} & {58.93} & {53.82} & {21.67} & {55.93} & {39.04} & {47.23} & {46.10}\\
                 DC-Net~\cite{tian2020robust} & {68.54} & {58.74} & {43.44} & {53.96} & {44.53} & {53.47} & {53.78}\\
                 Ours & {\textbf{79.59}} & \textbf{69.94} & {\textbf{75.24}} & {\textbf{62.57}} & {\textbf{56.98}} & {\textbf{68.02}} & {\textbf{68.72}}\\
                 \bottomrule
            \end{tabular}}}
            \label{table:quantitative_evaluation_sibp}
		\end{minipage}
\end{figure*}

\subsection{Validation of the Pose Selection Strategy}\label{subsec:validation_pose_selection}
We study the performance of our self-training method with different pose selection strategies. Specifically, we separately remove the appearance-based metric $d_a$ and geometry-based metric $d_g$ from the framework, and evaluate their performances on ROBI. 
\begin{figure*}[t!]
	\centering
		\begin{minipage}{0.98\textwidth}
			\centering
			\tabcaption{Quantitative evaluation of the proposed self-training method with different pose selection settings. $d_a$ denotes the appearance-based metric and $d_g$ denotes the geometry-based metric. LB denotes the lower bound of our model, which is trained with only synthetic data. UB denotes the upper bound of our model, which additionally uses pose labels of real data for training.}
            \resizebox{0.98\textwidth}{!}{%
            \renewcommand\arraystretch{0.95}
            \setlength{\tabcolsep}{1.3mm}{
            \begin{tabular}{cc|ccccccc|c}
                \toprule
                {$d_a$} & {$d_g$} & {Zigzag*} & \makecell[c]{Chrome\\Screw} & Gear & \makecell[c]{Eye\\Bolt} & \makecell[c]{Tube\\Fitting} & \makecell[c]{Din*\\Connector} & \makecell[c]{D-Sub*\\Connector} & {Mean}\\\hline
                \checkmark & {} & {38.54} & {78.00} & {91.29} & {85.09} & {95.52} & {22.83} & {12.06} & {60.48}\\
                {} & \checkmark & {43.10} & {\textbf{79.98}} & {95.20} & {87.84} & {96.91} & {21.61} & {13.95} & {62.66}\\
                \checkmark & \checkmark & {\textbf{45.83}} & {79.88} & {\textbf{97.81}} & {\textbf{89.21}} & {\textbf{97.45}} & {\textbf{24.21}} & {\textbf{14.62}} & {\textbf{64.14}}\\\hline
                \multicolumn{2}{c|}{LB} & {30.96} & {67.74} & {77.56} & {53.52} & {74.84} & {18.72} & {10.60} & {47.41}\\
                \multicolumn{2}{c|}{UB} & {56.90} & {93.71} & {98.90} & {96.12} & {99.46} & {44.16} & {33.86} & {74.73}\\
                \bottomrule
            \end{tabular}}}
            \label{table:2d3d_metric}
		\end{minipage}
		\vfill
		\begin{minipage}[h]{1.0\linewidth}
			\centering
			\begin{minipage}{0.95\linewidth}
                \includegraphics[width=1.0\linewidth]{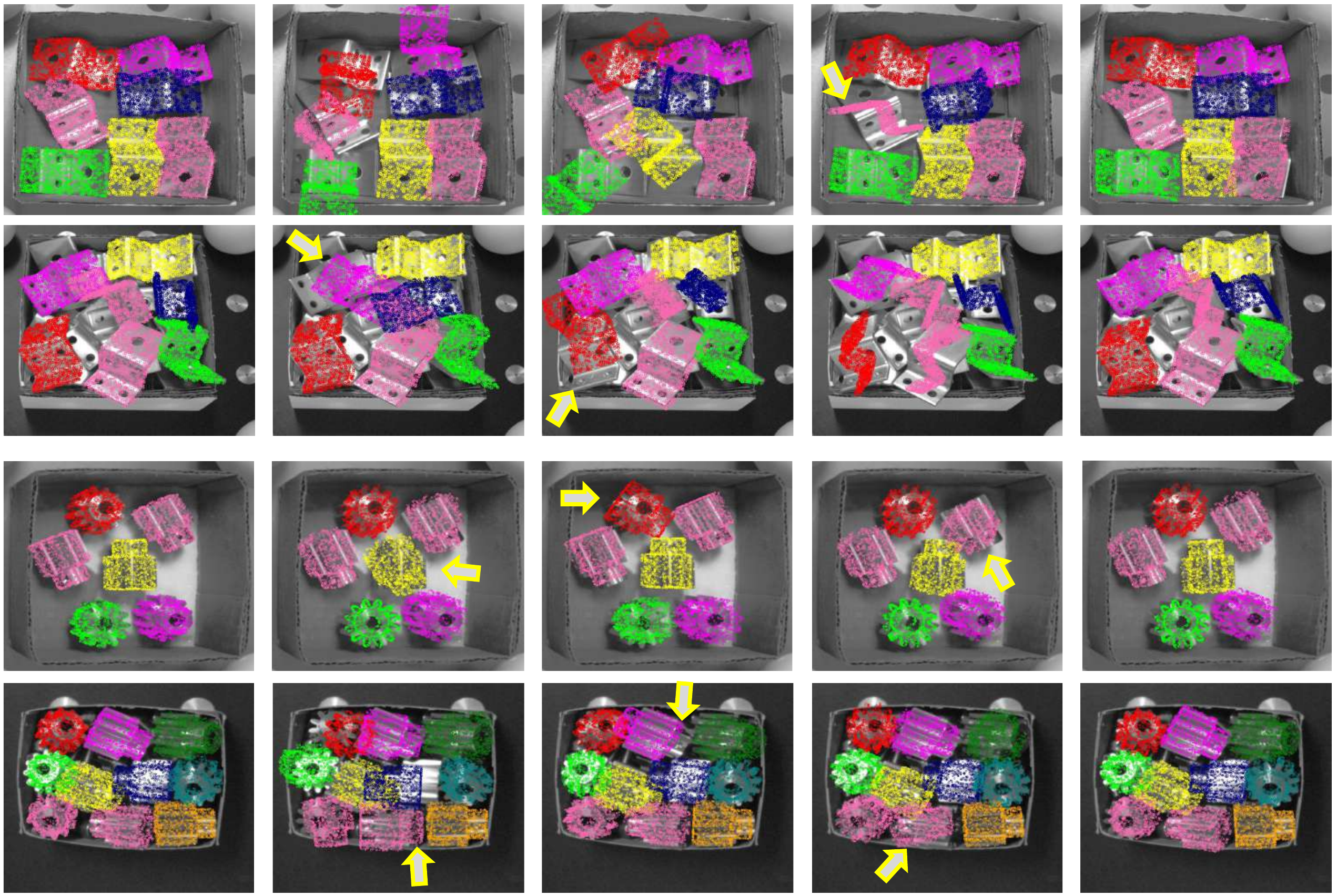}
            \end{minipage}
            \vfill
            \begin{minipage}{0.95\linewidth}
                \begin{minipage}{1.0\linewidth}
                    \begin{minipage}{0.192\textwidth}
                        \centering{GT}
                    \end{minipage}
                    \hfill
                    \begin{minipage}{0.192\textwidth}
                        \centering{MP-AAE~\cite{sundermeyer2020multi}}
                    \end{minipage}
                    \hfill
                    \begin{minipage}{0.192\textwidth}
                        \centering{AAE~\cite{sundermeyer2020augmented}}
                    \end{minipage}
                    \hfill
                    \begin{minipage}{0.192\textwidth}
                        \centering{DC-Net~\cite{tian2020robust}}
                    \end{minipage}
                    \hfill
                    \begin{minipage}{0.192\textwidth}
                        \centering{Ours}
                    \end{minipage}
                \end{minipage}
            \end{minipage}
            \caption{Qualitative comparison with state-of-the-arts. The 6D object pose estimation results are depicted by projecting the object model onto the image plane using the object pose and camera intrinsic parameters. DC-Net is our baseline model w/o our proposed self-training. Colored points denote the projected model~(best viewed in color).}
            \label{fig:qualitative_comparison}
		\end{minipage}
\end{figure*}
Tab.~\ref{table:2d3d_metric} reports the comparative results. Compared with the lower-bound model~(LB), both the proposed appearance metric and geometry metric help to significantly improve the pose accuracy by self-training. Among three different settings, using both appearance and geometry information for pose selection achieves the best pose accuracy. Removing $d_g$ results in a larger pose accuracy drop than removing $d_a$.
These experimental results indicate that merely assessing the pose label in 2D is not sufficient for self-training of 6D object pose. Using both 2D appearance and 3D geometry information provides the most reliable scheme for pose selection, and achieves the best pose accuracy.

\subsection{Self-training with Different Backbone Networks}\label{subsec:self_training_different_backbone}
In contrary to other self-supervised/self-training methods~\cite{deng2020self,gal2016dropout,pastore2021closer,wang2020self6d}, our proposed self-training framework can be easily applied to different networks for sim-to-real object pose estimation. In this section, we change the backbone network to DenseFusion~\cite{wang2019densefusion}, another popular RGB-D based object pose estimation network. Compared with DC-Net, DenseFusion is a two-stage method which is composed of an initial \textit{estimator} and a pose \textit{refiner}. We then use the same synthetic data and real data with previous experiments for self-training and evaluation. Tab.~\ref{table:densefusion} reports the experiment results. Even with a different backbone network, the proposed self-training method can consistently adapt the model to real data and significantly improve the pose accuracy on real data. 
\begin{figure*}[t]
    \centering
    \begin{minipage}{0.95\linewidth}
                \includegraphics[width=1.0\linewidth]{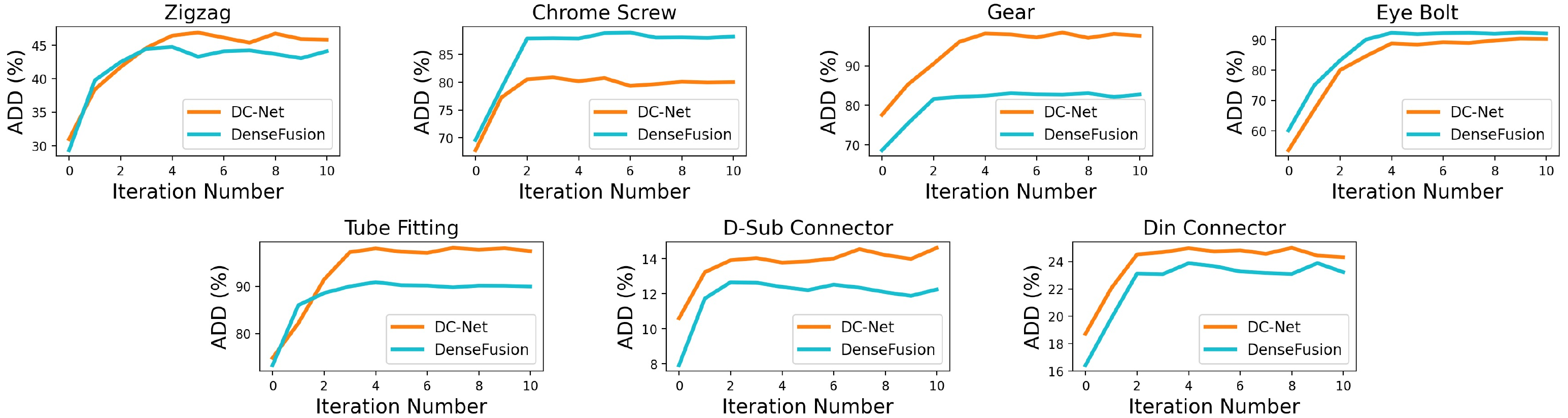}
            \end{minipage}
            \vfill
            \begin{minipage}{0.95\linewidth}
                \includegraphics[width=1.0\linewidth]{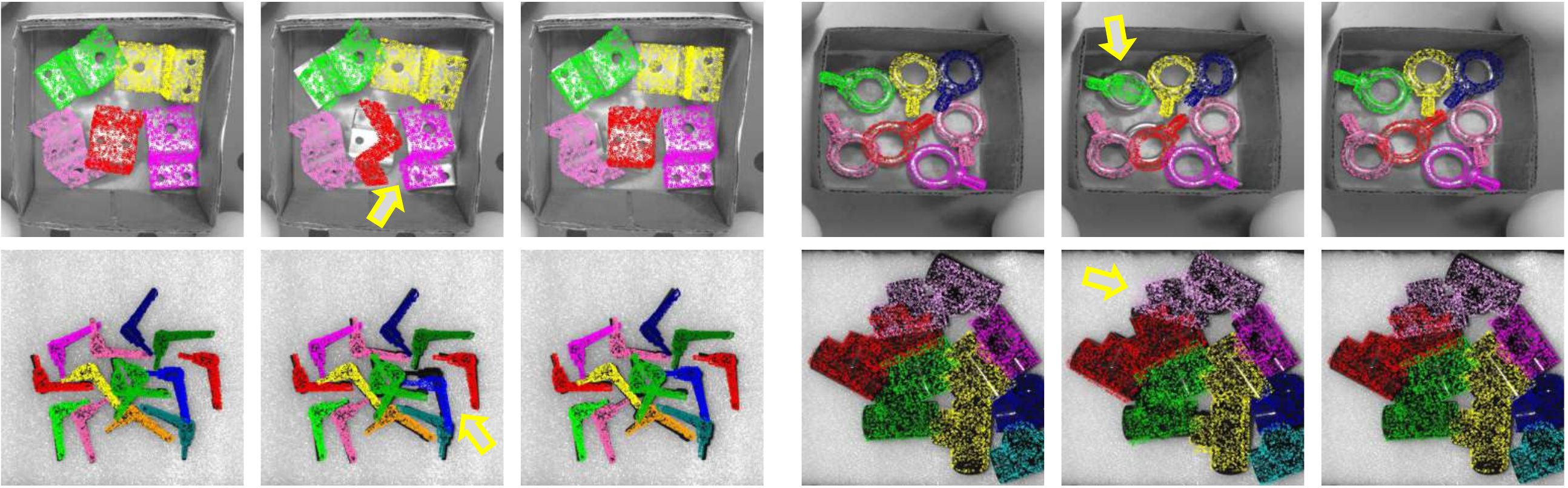}
            \end{minipage}
            \vfill
            \begin{minipage}{0.95\linewidth}
                \begin{minipage}{0.485\linewidth}
                    \begin{minipage}{0.32\linewidth}
                        \centering{(a)}
                    \end{minipage}
                    \hfill
                    \begin{minipage}{0.32\linewidth}
                        \centering{(b)}
                    \end{minipage}
                    \hfill
                    \begin{minipage}{0.32\linewidth}
                        \centering{(c)}
                    \end{minipage}
                \end{minipage}
                \hfill
                \begin{minipage}{0.485\linewidth}
                    \begin{minipage}{0.32\textwidth}
                        \centering{(d)}
                    \end{minipage}
                    \hfill
                    \begin{minipage}{0.32\textwidth}
                        \centering{(e)}
                    \end{minipage}
                    \hfill
                    \begin{minipage}{0.32\textwidth}
                        \centering{(f)}
                    \end{minipage}
                \end{minipage}
            \end{minipage}
            \figcaption{Object pose estimation results of our iterative self-training model. Top: The average recall of ADD(-S) metric on ROBI dataset with different number of iterations. Bottom: Qualitative results. (a)(d) are ground-truth results. (b)(e) are results of one-time self-training. (c)(f) are results of five-time iterative self-training.}
            \label{fig:iteration}
\end{figure*}

\subsection{Continuous Improvement with Iterative Self-training}\label{subsec:iterative_self_training}
In order to study the effect of iteration numbers on the final pose accuracy, we perform iterative self-training with a maximal number of 10 iterations, and compared the pose accuracy after each iteration. Fig.~\ref{fig:iteration} depicts the experiment results. In general, the pose error reduces as the number of self-training iterations increase. In most cases, the first iteration usually accounts for the most improvement of the pose accuracy. After that, the second and third iteration can continuously enhance the pose accuracy, with a relatively smaller improvement. After about 5 iterations, the performance slowly saturates without further improvement. Fig.~\ref{fig:iteration} further presents qualitative comparisons between one-time iteration model and five-time iteration model. With the increase of iterations, the pose estimation results get visually closer to the ground truth.

\begin{table*}[t]
    \centering
    \caption{Results of the proposed self-training method with a different backbone network~\cite{wang2019densefusion}. w/o ST and w/ ST denote models without and with proposed self-training.}
    \label{table:comparison_robi}
    \resizebox{0.98\textwidth}{!}{%
    \renewcommand\arraystretch{0.9}
    \setlength{\tabcolsep}{1.3mm}{
    \begin{tabular}{l|ccccccc|c}
    \toprule
    {} & {Zigzag*} & \makecell[c]{Chrome\\Screw} & Gear & \makecell[c]{Eye\\Bolt} & \makecell[c]{Tube\\Fitting} & \makecell[c]{Din*\\Connector} & \makecell[c]{D-Sub*\\Connector} & {Mean}\\\hline
    {w/o ST} & {29.29} & {69.61} & {68.59} & {60.07} & {73.29} & {16.42} & {7.92} & {46.46}\\
    {w/ ST} & {\textbf{44.46}} & {\textbf{87.87}} & {\textbf{82.73}} & \textbf{92.36} & \textbf{90.88} & \textbf{23.88} & \textbf{12.66} & \textbf{62.12}\\
    \bottomrule
    \end{tabular}}}
    \label{table:densefusion}
\end{table*}

\begin{table}[t]
	\centering
	\caption{Results of the robot bin-picking experiments. The \textit{left} picture illustrates the robot bin-picking environment. \textbf{Inst.} denotes the total number of objects to be grasped. \textbf{Tria.} denotes the total number of trials needed to access all of the objects.}
		\begin{minipage}{0.16\textwidth}
			\centering
			\includegraphics[width=1.0\linewidth]{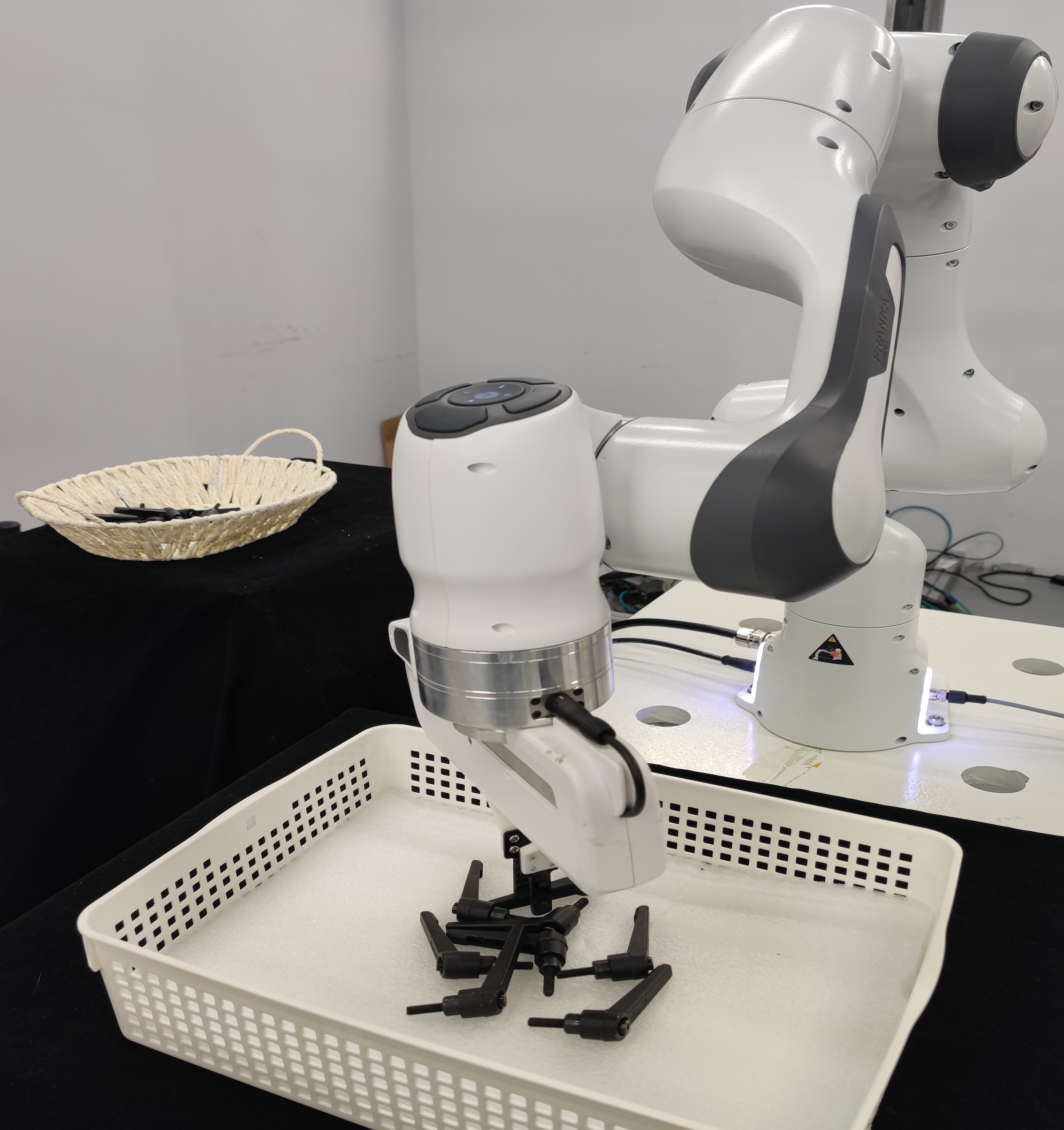}
		\end{minipage}
		\hfill
		\begin{minipage}[h]{0.8\textwidth}
			\centering
            \renewcommand\arraystretch{1.1}
            \setlength{\tabcolsep}{2.2mm}{
            \begin{tabular}{c|ccccc|c}
            \toprule
            \multirow{2}{*}{} & \multicolumn{5}{c|}{Inst. / Tria.} & \multirow{2}{*}{Mean}\\\cline{2-6}
            {} & 01 & 02 & 03 & 04 & 05 &{}\\\hline
            {w/o ST} & {14/24} & {14/20} & {14/18} & {14/22} & {14/19} & {67.96\%}\\
            {w/ ST} & {14/14} & {14/16} & {14/18} & {14/16} & {14/16} & {\textbf{87.50\%}}\\
            \bottomrule
            \end{tabular}}
		\end{minipage}
		\label{table:grasp}
\end{table}


\subsection{Effectiveness Demonstration on Robots}\label{subsec:robot_demonstration}
In order to further demonstrate the effectiveness of proposed self-training object pose estimation method, we deploy the trained pose estimation model on a Franka Emika Panda robot arm and conduct real-world bin-picking experiments on the \textit{`Handle'} object from SIBP. The grasp configuration is generated offline based on the object CAD model~\cite{fang2020graspnet}, and then projected to the camera coordinate system according to estimated object pose. This means that overall grasping success rate is highly reliant on pose accuracy. Tab.~\ref{table:grasp} presents the experiment results. Following self-training of the pose estimation network, the grasping success rate dramatically improves. These results demonstrate the effectiveness of our proposed self-training method for practical industrial bin-picking.
\section{Conclusion}\label{sec:conclusion}
With the goal of achieving accurate 6D object pose estimation in real-world scenes, we have presented a sim-to-real method that first trains a 6D pose estimation on high-fidelity simulation data, and then performs our iterative self-training method. Our method provides an efficient solution for pose estimation where labeled data is hard~(or even impossible) to collect. Extensive results demonstrate that our approach can significantly improve the predicted pose quality, with great potential to be applied to industrial robotic bin-picking scenarios.

Currently, our method assumes access to object models. Although this is acceptable in the industrial bin-picking domain and is indeed common in the pose estimation literature, it would nonetheless be interesting to extend this iterative self-training method to pose estimation of previously unseen objects~\cite{chen2021sgpa,wang2021category}. Our method exploits object mask predicted by a segmentation network trained with synthetic data. It would be meaningful to extend our iterative self-training method to joint instance segmentation and object pose estimation.
\\
\\
\noindent\textbf{Acknowledgement.}
The work was supported by the Hong Kong Centre for Logistics Robotics.

\bibliographystyle{splncs04}
\bibliography{egbib}
\clearpage

\section*{Appendix}\label{sec:appendix}
\noindent In this supplementary material, we provide additional contents that are not included in the main paper due to the space limit:
\begin{itemize}
    \item Training details of the proposed self-training method for sim-to-real 6D object pose estimation~(Section A)
    \item More details about ROBI dataset and SIBP dataset~(Section B)
    \item Results with different amount of real data for self-training~(Section C)
    \item More qualitative comparison results~(Section D)
\end{itemize}

\subsection*{A. Details of Network Output and Loss Function}
Our proposed iterative self-training method for sim-to-real object pose estimation is designed to be scalable and network-agnostic. The specific output format for representing the 6D object pose and the loss function are dependent on the architecture of the adopted object pose estimation backbone network. In our experiments, we mainly take DC-Net~\cite{tian2020robust} as the backbone network to test the performance of the proposed self-training method. In this case, the network output and the loss function used for self-training are similar with the ones proposed in~\cite{tian2020robust}.

For the rotation, we use discrete-continuous formulation for regressing rotation. Specifically, based on the icosahedral group, we generate 60 rotation anchors to uniformly sample the whole $SO(3)$ space. For each rotation anchor, the network would predict a rotation deviation in the form of quaternion and an uncertainty value $\sigma$ to indicate the confidence of each rotation anchor. For the translation, the network would predict a unit vector $v'$ for each input point that represents the direction from the point to the object center. Then, the translation is estimated based on a RANSAC-based voting. The loss function used for rotation estimation is based on the ShapeMatch-Loss~\cite{xiang2017posecnn}. For symmetric objects, it is defined as:
\begin{equation}
    L=\frac{1}{M}\sum_{x_1\in\mathcal{M}}\min_{x_2\in\mathcal{M}}\|\widetilde{R}x_1-R'x_2\|_2,
\end{equation}
where $M$ denotes the total number of points of the object CAD model. $\widetilde{R}$ and $R'$ denote rotations correspond to the pseudo label and the network output respectively. For asymmetric objects, the loss function is defined as:
\begin{equation}
    L=\frac{1}{M}\sum_{x\in\mathcal{M}}\|\widetilde{R}x-R'x\|_2.
\end{equation}
The specific probabilistic loss used for rotation estimation that considers the uncertainty value predicted by the network is defined as:
\begin{equation}
    L_R=\sum_{i}=\text{ln}\sigma_i+\frac{L_i}{d\times \sigma_i},
\end{equation}
where $d$ denotes the diameter of the object. In addition, we use smooth L1 loss for translation estimation:
\begin{equation}
    L_t=\left\{
    \begin{aligned}
        & \sum_{i}\frac{1}{M}\sum_{j}0.5\times\|\widetilde{v_{ij}}-v_{ij}'\|_2^2, &\|\widetilde{v_{ij}}-v_{ij}'\| < 1.0\\
        & \sum_{i}\frac{1}{M}\sum_{j}\|\widetilde{v_{ij}}-v_{ij}'\| - 0.5, &\text{otherwise}\\
    \end{aligned}
    \right.
\end{equation}
where $M$ denotes the total number of points of the object CAD model. $\widetilde{v}$ and $v'$ denote unit vectors computed from the pseudo label and the network prediction respectively. We integrate loss functions for rotation estimation and translation estimation as in~\cite{tian2020robust} for the network training.
\begin{figure*}
    \centering
        \begin{minipage}[h]{1.0\linewidth}
            \begin{minipage}{1.0\linewidth}
                \includegraphics[width=1.0\linewidth]{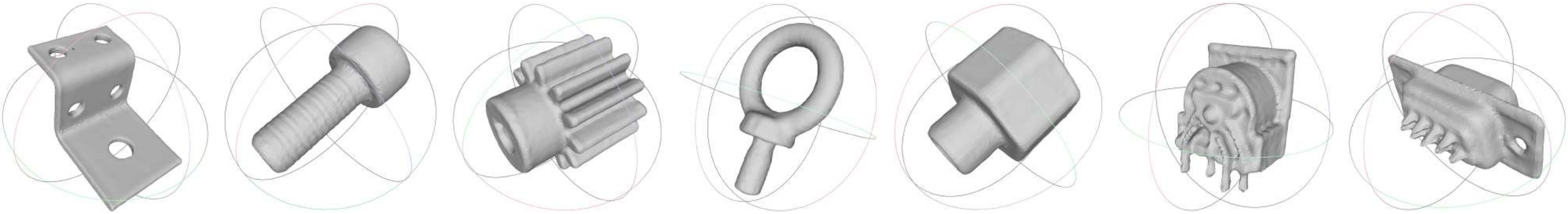}
            \end{minipage}
            \vfill
            \begin{minipage}{1.0\linewidth}
                \begin{minipage}{1.0\linewidth}
                    \begin{minipage}{0.135\textwidth}
                        \centering{Zigzag}
                    \end{minipage}
                    \hfill
                    \begin{minipage}{0.135\textwidth}
                        \centering{Chrome Screw}
                    \end{minipage}
                    \hfill
                    \begin{minipage}{0.135\textwidth}
                        \centering{Gear}
                    \end{minipage}
                    \hfill
                    \begin{minipage}{0.135\textwidth}
                        \centering{Eye Bolt}
                    \end{minipage}
                    \hfill
                    \begin{minipage}{0.135\textwidth}
                        \centering{Tube Fitting}
                    \end{minipage}
                    \hfill
                    \begin{minipage}{0.135\textwidth}
                        \centering{Din Connector}
                    \end{minipage}
                    \hfill
                    \begin{minipage}{0.135\textwidth}
                        \centering{D-Sub Connector}
                    \end{minipage}
                \end{minipage}
            \end{minipage}
            \caption{Seven object models provided by ROBI~\cite{yang2021robi}. Zigzag, Din Connector, and D-Sub Connector are three asymmetric objects, and the remains are symmetric objects. Din Connector and D-Sub Connector are composed of two different materials.}
            \label{fig:robi_model}
		\end{minipage}
		\vfill
		\begin{minipage}{1.0\linewidth}
            \includegraphics[width=1.0\linewidth]{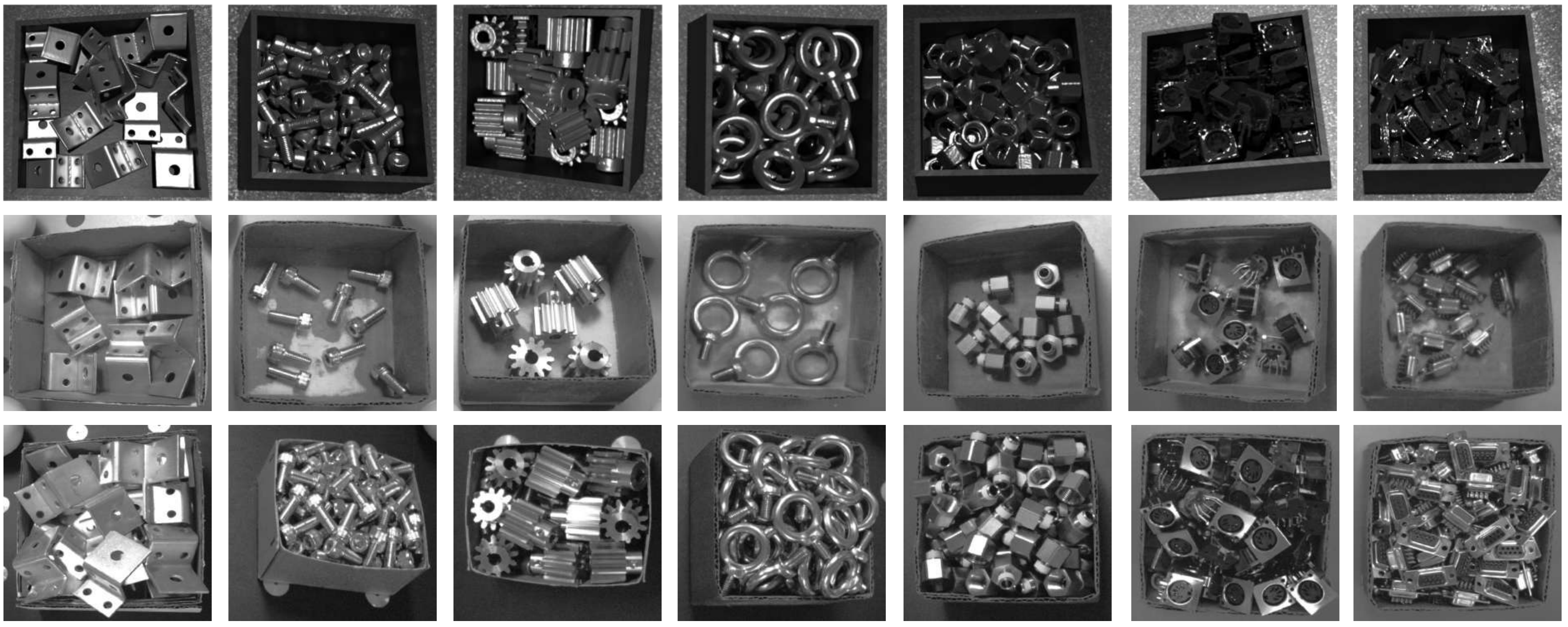}
            \caption{Examples of ROBI dataset. The first row presents the synthetic data. The second and third rows show real data in low-bin and full-bin scenarios.}
            \label{fig:robi_data_example}
        \end{minipage}
        \vfill
        \begin{minipage}{1.0\linewidth}
            \begin{minipage}{1.0\linewidth}
                \includegraphics[width=1.0\linewidth]{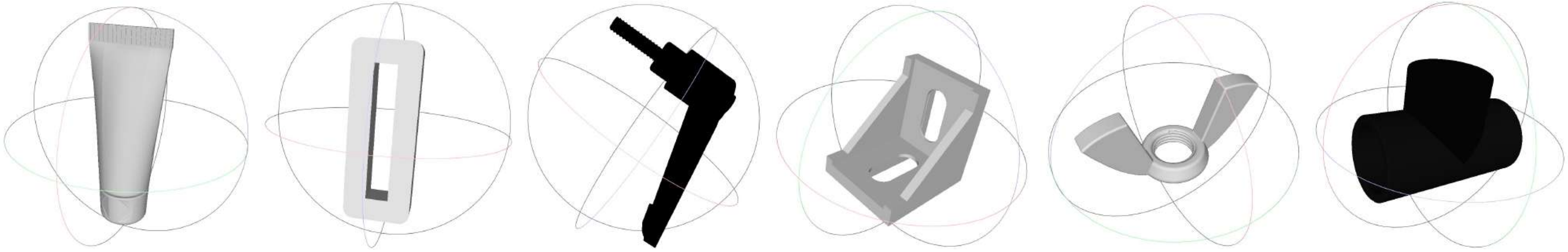}
            \end{minipage}
            \vfill
            \begin{minipage}{1.0\linewidth}
                \begin{minipage}{1.0\linewidth}
                    \begin{minipage}{0.16\textwidth}
                        \centering{Cosmetic}
                    \end{minipage}
                    \hfill
                    \begin{minipage}{0.16\textwidth}
                        \centering{Flake}
                    \end{minipage}
                    \hfill
                    \begin{minipage}{0.16\textwidth}
                        \centering{Handle}
                    \end{minipage}
                    \hfill
                    \begin{minipage}{0.16\textwidth}
                        \centering{Corner}
                    \end{minipage}
                    \hfill
                    \begin{minipage}{0.16\textwidth}
                        \centering{Screw Head}
                    \end{minipage}
                    \hfill
                    \begin{minipage}{0.16\textwidth}
                        \centering{T-Shape Connector}
                    \end{minipage}
                \end{minipage}
            \end{minipage}
            \caption{Six objects used in SIBP. Handle is an asymmetric object, and remains are symmetric objects. }
            \label{fig:SIBP_model}
        \end{minipage}
        \vfill
        \begin{minipage}{1.0\linewidth}
            \begin{minipage}{1.0\linewidth}
                \includegraphics[width=1.0\linewidth]{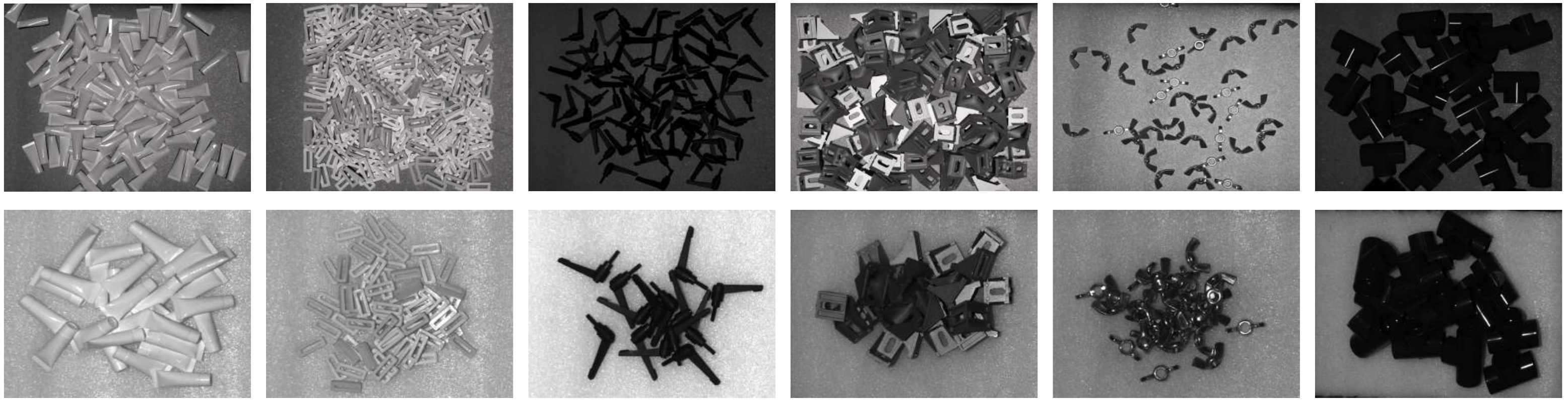}
            \end{minipage}
            \caption{Examples of SIBP dataset. The first row presents the synthetic data. The second row shows the real data.}
            \label{fig:sibp_data_example}
        \end{minipage}
\end{figure*}

\subsection*{B. Detailed Information about Experiment Dataset}
As mentioned in the main paper, we extend ROBI with $7000$ synthetic scenes based on the provided CAD models shown in Fig.~\ref{fig:robi_model}. We further build the SIBP dataset. It provides both synthetic and real data for $6$ objects with different materials and sizes. When synthesizing virtual data for both ROBI and SIBP, we follow~\cite{denninger2020blenderproc,xianzhi2022s2r} and leverage multiple strategies to narrow the sim-to-real gap: (a) add realism surface texture based on the provided texture-less CAD model; (b) simulate virtual cluttered scene under the physical constraints; (c) employ ray-tracing rendering engine in Blender~\cite{blender} to generate photo-realistic images; (d) use copy-paste strategy to randomize background. Fig.~\ref{fig:robi_data_example} presents some examples of our extended ROBI dataset.

Fig.~\ref{fig:SIBP_model} depicts $6$ objects that are adopted in constructing the SIBP dataset. We use a fixed industrial stereo camera to collect the required RGB-D data. To make the collected real data of SIBP closer to the practical bin-picking scenario, we follow the scheme in~\cite{kleeberger2019large} to collect RGB-D data for each object. We first put tens of objects in the scene. Then, we carefully remove the objects from the scene one by one and keep the remaining objects unchanged. Fig.~\ref{fig:sibp_data_example} depicts some examples of the SIBP dataset. Tab.~\ref{tab:statistics_sibp} further provides detailed statistics information of SIBP.

\begin{table*}[t]
    \centering
    \caption{Statistics information of SIBP dataset.}
    \renewcommand\arraystretch{1.1}
    \setlength{\tabcolsep}{1.5mm}{
    \begin{tabular}{l|cccccc}
    \toprule
    {} & {Cosmetic} & {Flake} & {Handle} & {Corner} & \makecell[c]{Screw\\Head} & \makecell[c]{T-Shape\\Connector}\\\hline
    {Diameter~(mm)} & {67.9} & {34.5} & {77.4} & {66.6} & {54.5} & {108.4}\\
    {Surface material} & {plastic} & {plastic} & {metallic} & {alloyed} & {metallic} & {plastic}\\
    {Geometric symmetry} & {yes} & {yes} & {no} & {yes} & {yes} & {yes}\\
    {Synthetic data} & {1000} & {1000} & {1000} & {1000} & {1000} & {1000}\\
    {Real data} & {529} & {588} & {418} & {480} & {449} & {279}\\
    \bottomrule
    \end{tabular}}
    \label{tab:statistics_sibp}
\end{table*}
\begin{figure*}
    \centering
    \begin{minipage}{1.0\linewidth}
        \includegraphics[width=1.0\linewidth]{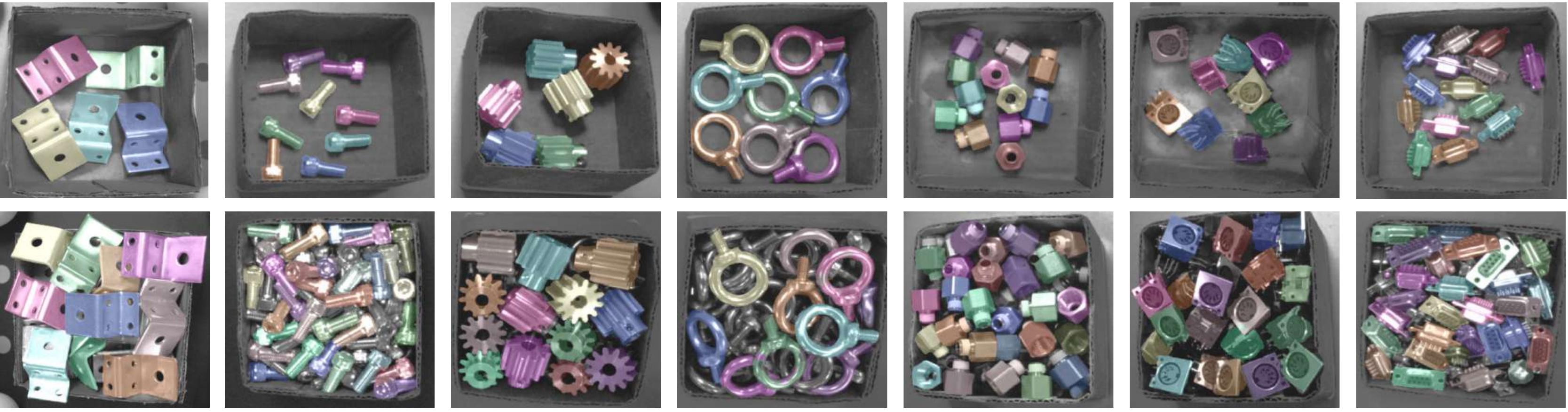}\vspace{-4mm}
        \caption{Segmentation results on ROBI with a Mask-RCNN trained on synthetic data.}
        \label{fig:ROBI_segmentation}
    \end{minipage}
    \vfill
    \vspace{3mm}
    \begin{minipage}{1.0\linewidth}
        \includegraphics[width=1.0\linewidth]{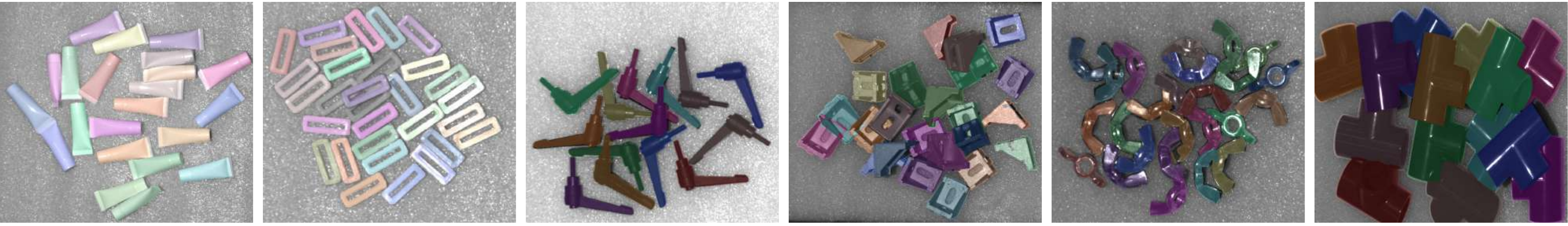}\vspace{-4mm}
        \caption{Segmentation results on SIBP with a Mask-RCNN trained on synthetic data.}\vspace{-3mm}
        \label{fig:SIBP_segmentation}
    \end{minipage}
\end{figure*}

In addition, before object pose estimation, we trained a Mask-RCNN~\cite{he2017mask} with the synthetic data for instance segmentation. Thanks to photo-realistic and physically plausible rendering techniques, as well as many off-the-shelf but effective 2D augmentation techniques~\cite{ghiasi2021simple}, the predicted instance masks on real data could be very accurate to be used for self-training. Fig.~\ref{fig:ROBI_segmentation} and Fig.~\ref{fig:SIBP_segmentation} depict some qualitative examples for instance segmentation.

\subsection*{C. Self-training with Different Amounts of Real Data}
\begin{figure*}[h]
    \centering
    \begin{minipage}{0.45\linewidth}
        \includegraphics[width=1.0\linewidth]{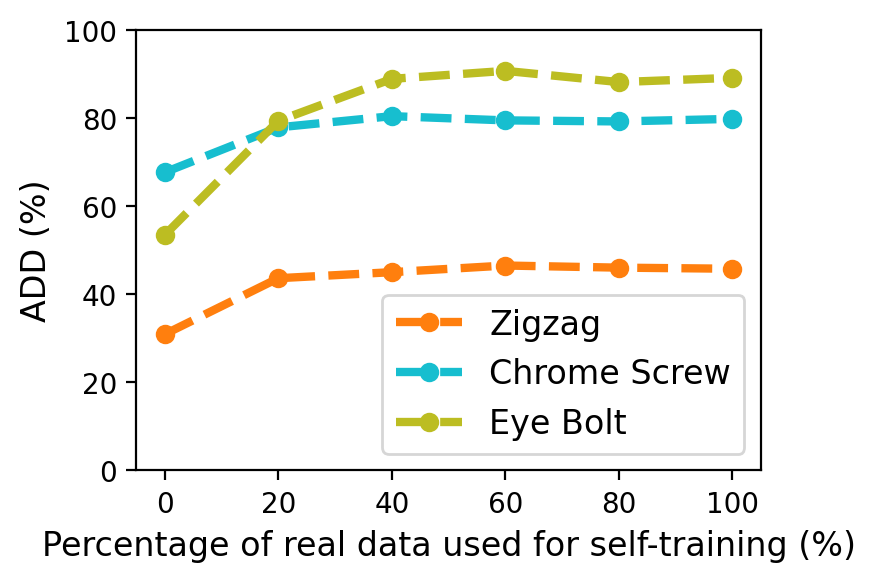}
    \end{minipage}\vspace{-3mm}
    \caption{Self-training results with different amount of real data for self-training. }\vspace{-4mm}
    \label{fig:results_different_data_amount}
\end{figure*}
We further studied the self-training performance with different amounts of real data on three objects of ROBI. We trained the object pose estimation network on $20\%$, $40\%$, $60\%$, $80\%$ and $100\%$ of training data with the proposed self-training method, and tested the resulting model on the same testing dataset. Fig.~\ref{fig:results_different_data_amount} presents the experiment results. With only $20\%$ real data, the self-training model has significantly outperformed the model trained with only synthetic data. With about $40\%$ real data, the model achieved comparable performance with the model trained with $100\%$ real data. These results demonstrate the data efficiency of our proposed self-training method for sim-to-real object pose estimation in industrial bin-picking scenario.

\subsection*{D. More Qualitative Results}

\begin{figure*}[tbh]
    \centering
    \begin{minipage}{0.95\linewidth}
        \includegraphics[width=1.0\linewidth]{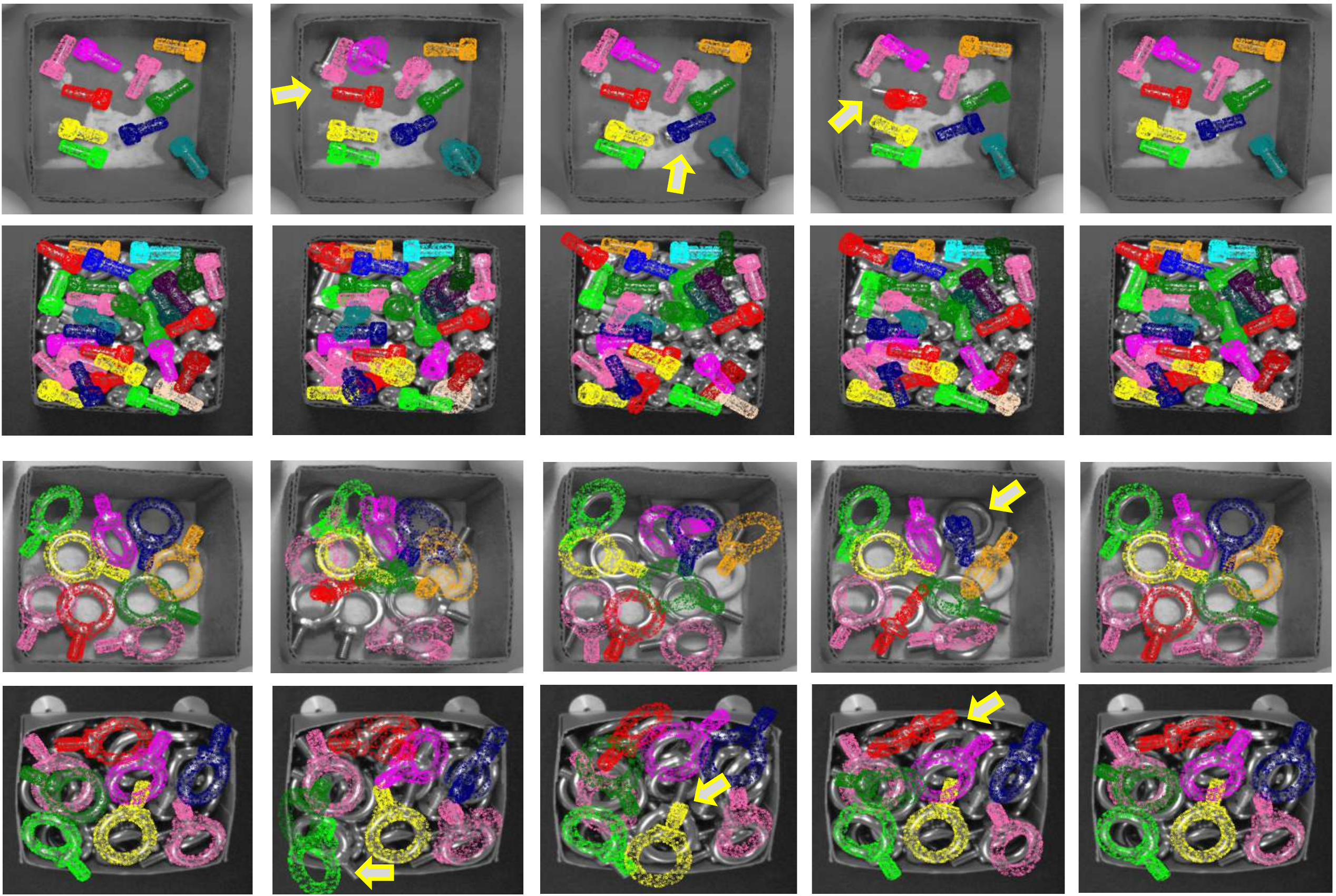}
    \end{minipage}
    \vfill
    \vspace{2mm}
    \begin{minipage}{0.95\linewidth}
        \begin{minipage}{1.0\linewidth}
            \begin{minipage}{0.192\textwidth}
                \centering{GT}
            \end{minipage}
            \hfill
            \begin{minipage}{0.192\textwidth}
                \centering{MP-AAE~\cite{sundermeyer2020multi}}
            \end{minipage}
            \hfill
            \begin{minipage}{0.192\textwidth}
                \centering{AAE~\cite{sundermeyer2020augmented}}
            \end{minipage}
            \hfill
            \begin{minipage}{0.192\textwidth}
                \centering{DC-Net~\cite{tian2020robust}}
            \end{minipage}
            \hfill
            \begin{minipage}{0.192\textwidth}
                \centering{Ours}
            \end{minipage}
        \end{minipage}
    \end{minipage}
    \vspace{-2mm}
    \caption{More qualitative comparisons with state-of-the-art methods on ROBI Dataset.}\vspace{-7mm}
    \label{fig:more_qualitative_comparison1}
\end{figure*}

\begin{figure*}[t]
    \centering
     \begin{minipage}{0.95\linewidth}
        \includegraphics[width=1.0\linewidth]{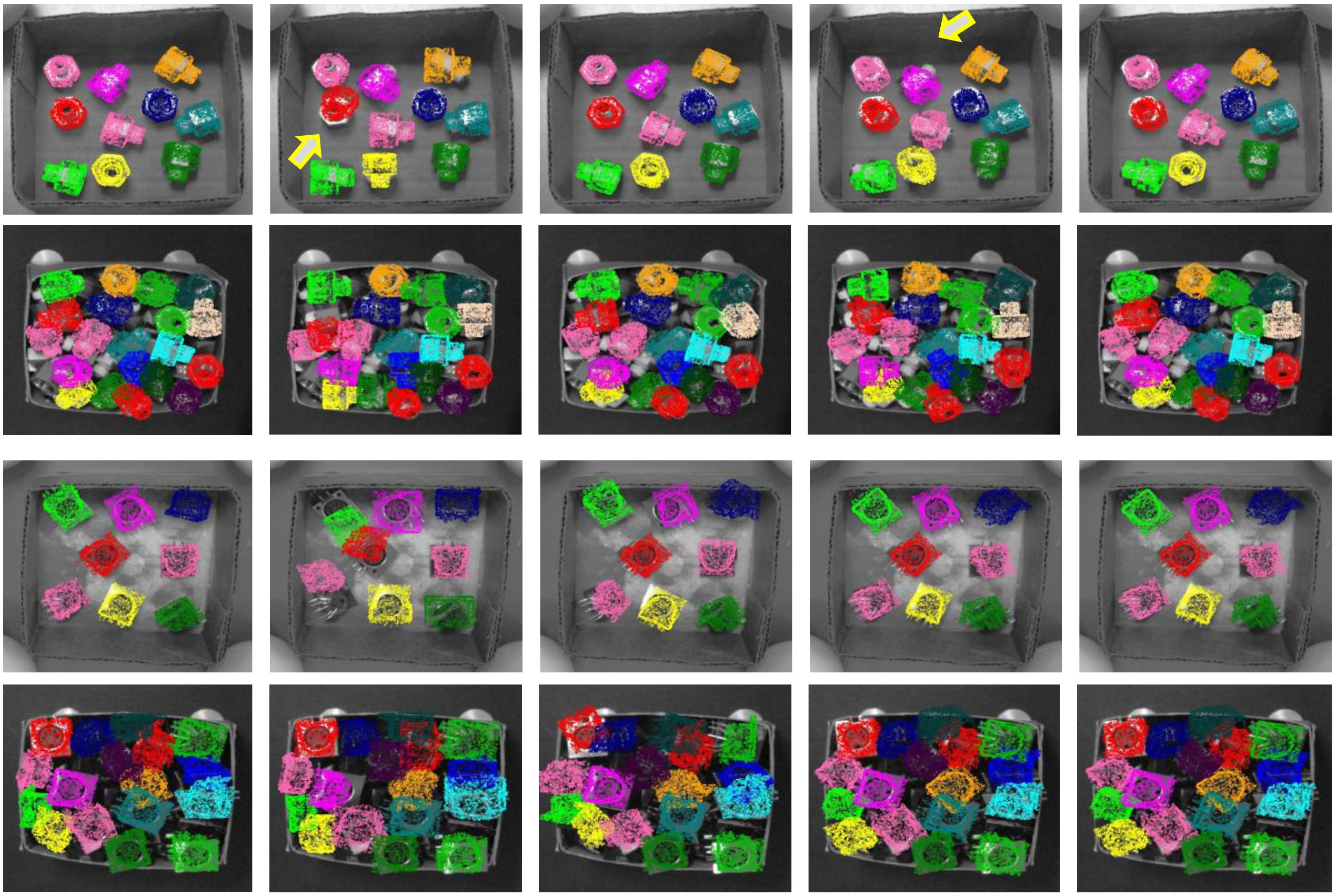}
    \end{minipage}
    \vfill
    \vspace{2mm}
    \begin{minipage}{0.95\linewidth}
        \includegraphics[width=1.0\linewidth]{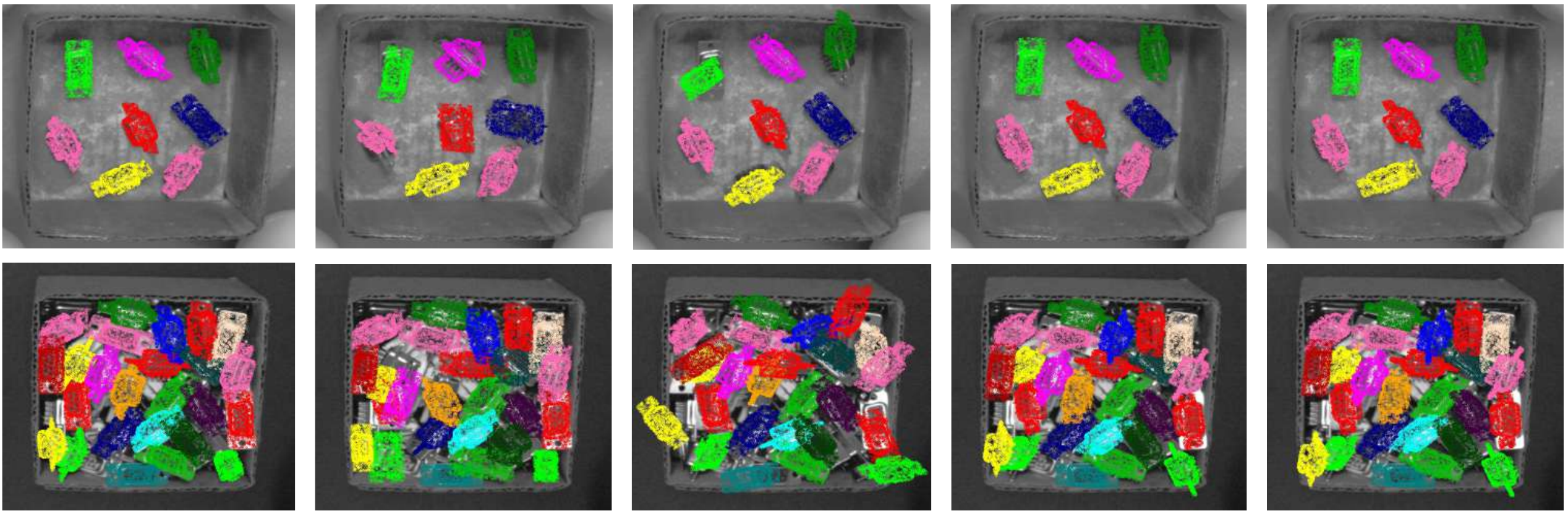}
    \end{minipage}
    \vfill
    \vspace{2mm}
    \begin{minipage}{0.95\linewidth}
        \begin{minipage}{1.0\linewidth}
            \begin{minipage}{0.192\textwidth}
                \centering{GT}
            \end{minipage}
            \hfill
            \begin{minipage}{0.192\textwidth}
                \centering{MP-AAE~\cite{sundermeyer2020multi}}
            \end{minipage}
            \hfill
            \begin{minipage}{0.192\textwidth}
                \centering{AAE~\cite{sundermeyer2020augmented}}
            \end{minipage}
            \hfill
            \begin{minipage}{0.192\textwidth}
                \centering{DC-Net~\cite{tian2020robust}}
            \end{minipage}
            \hfill
            \begin{minipage}{0.192\textwidth}
                \centering{Ours}
            \end{minipage}
        \end{minipage}
    \end{minipage}
    \vspace{-2mm}
    \caption{More qualitative comparisons with state-of-the-art methods on ROBI Dataset.}
    \label{fig:more_qualitative_comparison2}
\end{figure*}

\begin{figure*}[tbh]
    \centering
    \begin{minipage}{0.95\linewidth}
        \includegraphics[width=1.0\linewidth]{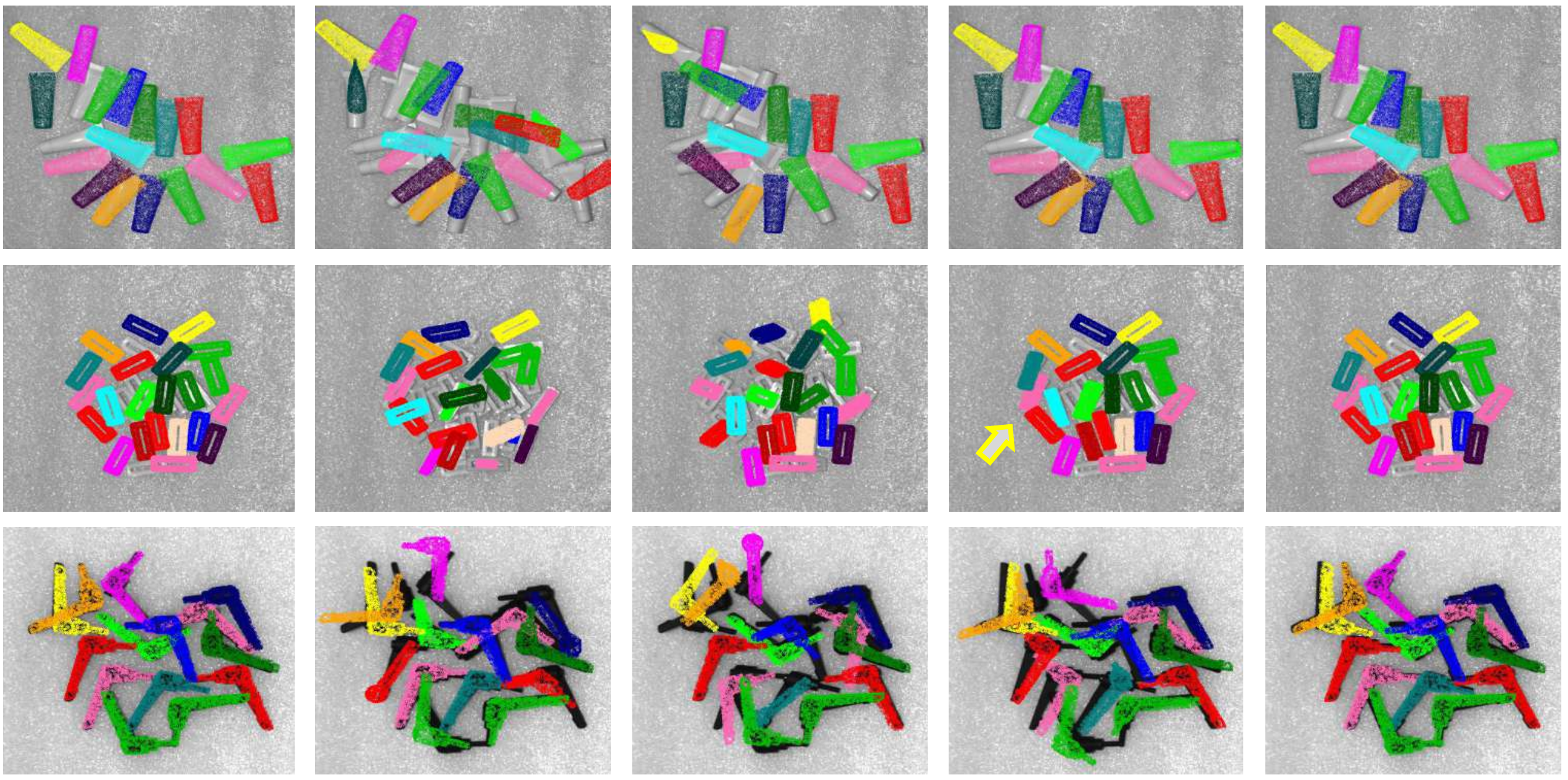}
    \end{minipage}
    \vfill
    \begin{minipage}{0.95\linewidth}
        \includegraphics[width=1.0\linewidth]{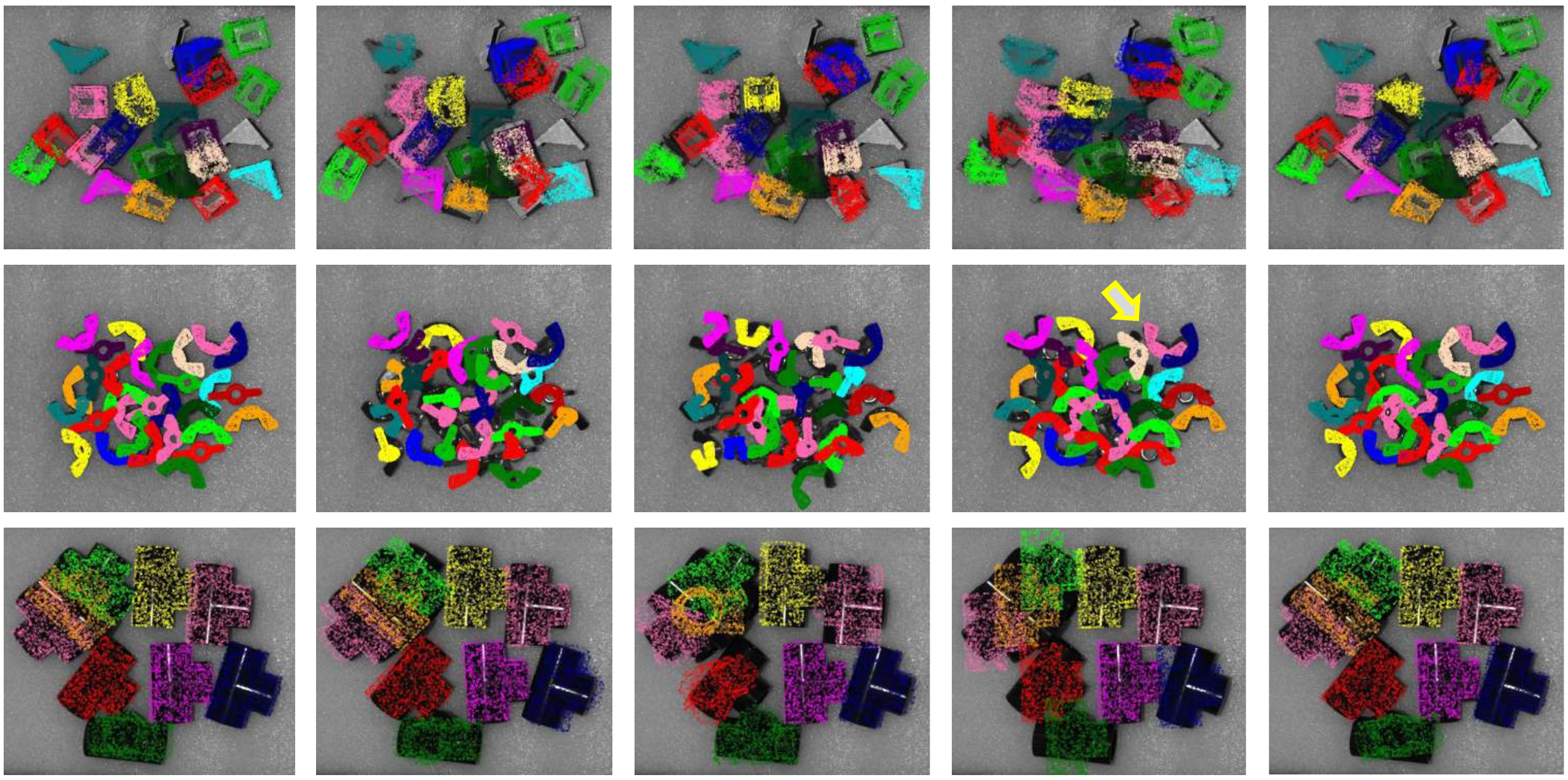}
    \end{minipage}
    \vfill
    \vspace{2mm}
    \begin{minipage}{0.95\linewidth}
        \begin{minipage}{1.0\linewidth}
            \begin{minipage}{0.192\textwidth}
                \centering{GT}
            \end{minipage}
            \hfill
            \begin{minipage}{0.192\textwidth}
                \centering{MP-AAE~\cite{sundermeyer2020multi}}
            \end{minipage}
            \hfill
            \begin{minipage}{0.192\textwidth}
                \centering{AAE~\cite{sundermeyer2020augmented}}
            \end{minipage}
            \hfill
            \begin{minipage}{0.192\textwidth}
                \centering{DC-Net~\cite{tian2020robust}}
            \end{minipage}
            \hfill
            \begin{minipage}{0.192\textwidth}
                \centering{Ours}
            \end{minipage}
        \end{minipage}
    \end{minipage}
    \vspace{-2mm}
    \caption{More qualitative comparisons with state-of-the-art methods on SIBP Dataset.}\vspace{-7mm}
    \label{fig:more_qualitative_comparison4}
\end{figure*}
\end{document}